\pdfoutput=1
\documentclass[11pt]{article}

\usepackage[preprint]{acl}

\usepackage{times}
\usepackage{latexsym}

\usepackage[T1]{fontenc}
\usepackage[utf8]{inputenc}

\usepackage{microtype}

\usepackage{inconsolata}

\usepackage{graphicx}
\usepackage{courier}
\usepackage{amsmath}
\usepackage{amsfonts}
\usepackage{array}
\usepackage{subfig}
\usepackage{CJKutf8}
\usepackage[most]{tcolorbox}
\usepackage{tcolorbox} 
\usepackage{caption}
\usepackage{hyperref}
\usepackage{algorithm}
\usepackage{algorithmic}
\usepackage{adjustbox}
\usepackage{float}
\usepackage{booktabs}
\usepackage{enumitem}
\usepackage{xcolor}  
\usepackage{colortbl} 

\definecolor{myPurple}{rgb}{0.4706, 0, 0.3922}

\newcommand{\ours}[0]{\texttt{DLPO}\ }
\definecolor{tkcolor}{RGB}{224,223,255}
\newtcolorbox{takeaways}[1][]{
	width=\columnwidth,
	colback = tkcolor, 
	colframe = tkcolor, 
	boxsep=0pt,left=10pt,right=10pt,top=2pt,bottom=2pt,
	fontupper=\linespread{0.9}\selectfont,
	title=#1}
\newtcolorbox{mybox}[2][]{
    width=\columnwidth,
    colback = gray!8, 
    colframe = gray!8, 
    boxsep=0pt,left=10pt,right=10pt,top=0pt,bottom=0pt,
    fontupper=\linespread{0.9}\selectfont,
    title=#2,#1}

\title{\ours: Towards a Robust, Efficient, and Generalizable Prompt Optimization Framework from a Deep-Learning Perspective}

\author{Dengyun Peng$^{\spadesuit}$\thanks{\ \ Equal Contribution} \quad Yuhang Zhou$^{\spadesuit \diamondsuit}$\footnotemark[1] \quad Qiguang Chen$^{\spadesuit}$\footnotemark[1]  \quad Jinhao Liu$^{\spadesuit}$ \\
\textbf{Jingjing Chen$^{\diamondsuit}$} \quad 	\textbf{Libo Qin$^{\clubsuit}$}\thanks{\ \ Corresponding Author} \\
	$^{\spadesuit}$ Research Center for Social Computing and Interactive Robotics \\
	$^{\spadesuit}$ Harbin Institute of Technology, China\\
	$^{\clubsuit}$  School of Computer Science and Engineering, Central South University, China \\
  $^{\diamondsuit}$ Fudan University, China\\
	\texttt{\{dypeng,qgchen\}@ir.hit.edu.cn}, \texttt{ralph.yh.zhou@gmail.com}\\
  \texttt{lbqin@csu.edu.cn}\\}

\begin{document}
\maketitle

\begin{abstract}
Large Language Models (LLMs) have achieved remarkable success across diverse tasks, largely driven by well-designed prompts. However, crafting and selecting such prompts often requires considerable human effort, significantly limiting its scalability. To mitigate this, recent studies have explored automated prompt optimization as a promising solution. Despite these efforts, existing methods still face critical challenges in \textit{\textbf{robustness, efficiency, and generalization}}.
To systematically address these challenges, we first conduct an empirical analysis to identify the limitations of current reflection-based prompt optimization paradigm.
Building on these insights, we propose 7 innovative approaches inspired by traditional deep learning paradigms for prompt optimization (\texttt{DLPO}), seamlessly integrating these concepts into text-based gradient optimization. 
Through these advancements, we progressively tackle the aforementioned challenges and validate our methods through extensive experimentation.
We hope our study not only provides valuable guidance for future research but also offers a comprehensive understanding of the challenges and potential solutions in prompt optimization.
Our code is available at \url{https://github.com/sfasfaffa/DLPO
}.
\end{abstract}
\section{Introduction}

\begin{figure}[t!]
    \centering
    \subfloat{
        \includegraphics[width=1\linewidth]{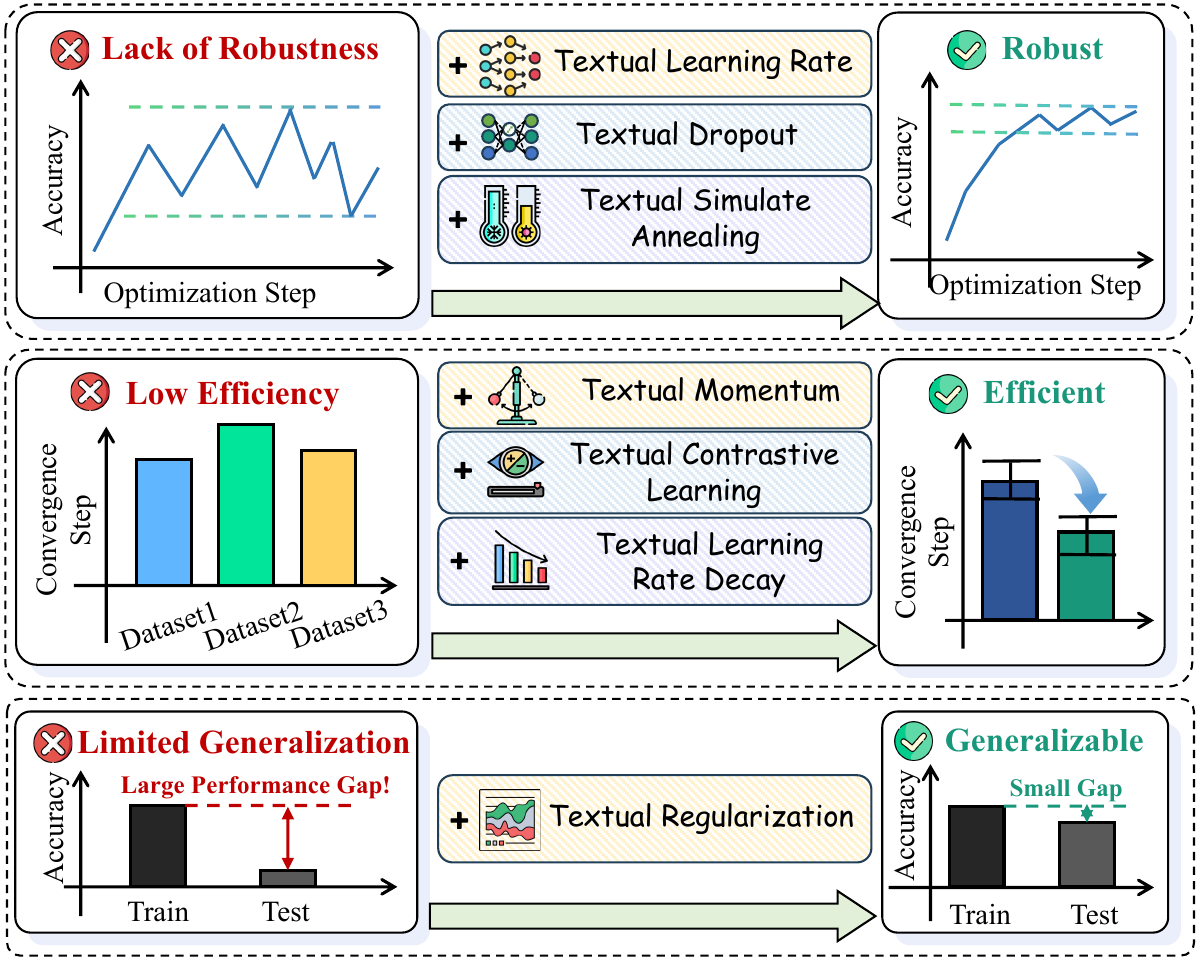}
    }
    \caption{Comparison between traditional reflection-based prompt optimization methods and \texttt{DLPO}, which incorporates \textbf{7 innovative approaches} to progressively enhance the robustness, efficiency, and generalizability of prompt optimization.}
    \label{fig:method-comparison}
\end{figure}

Large Language Models (LLMs) have demonstrated remarkable achievements across diverse applications~\citep{brown2020language, ouyang2022training, zhao2023survey,qin2024large}. Notably, the performance of LLMs highly relies on well-crafted prompts~\citep{zhu2023promptbench,li2024evaluating}.
Given their sensitivity to prompts, selecting optimal prompts is crucial for maximizing their performance in downstream tasks, a process known as Prompt Engineering \citep{reynolds2021prompt,sahoo2024systematic}, which requires substantial human effort~\citep{zamfirescu2023johnny}.
To address this, researchers have developed automated prompt optimization (PO) techniques, such as reinforcement learning-based methods~\citep{deng2022rlprompt, zhang2022tempera, diao2022black},
search-based methods~\citep{prasad2022grips,pryzant2023automatic,zhou2022large}, all aiming to improve efficiency and stability.
Among these methods, a dominant paradigm leverages LLMs' reflection capabilities, optimizing prompts based on feedback from external environments~\citep{wang2023promptagent,tang2024unleashing}.
Following \citet{pryzant2023automatic,yuksekgonul2024textgrad}, this paradigm draws inspiration from neural network training, reinterpreting key concepts: treating prompts as model parameters, output metrics as loss, and LLM reflection as gradients within textual expression. This framework has proven to be a powerful and widely adopted tool for various downstream tasks~\citep{zhou2024symbolic,manas2024improving,du2024ipo}.

Despite significant advancements in automated PO, our preliminary exploration reveals that existing methods encounter practical limitations that hinder their widespread adoption. These challenges can be categorized into three key issues, as illustrated in Fig.\ref{fig:method-comparison}:
(1) \textit{Lack of Robustness.} Current update methods often exhibit substantial oscillations and instability during the update process.
(2) \textit{Low Efficiency.} Existing PO strategies require massive iterations to achieve ideal prompts.
(3) \textit{Limited Generalizability.} While existing methods perform well on in-domain tasks present in training data, they struggle with distributionally out-of-domain tasks, restricting their applicability in real-world scenarios.

To address this, we first perform a systematic review of the prevailing reflection-based PO paradigm to identify its limitations.
Inspired by traditional deep learning approaches, we propose novel techniques for \texttt{DLPO} to enhance its PO capabilities. 
Traditional machine learning (ML) paradigm considers prompts as a single module, and relies on monotonic and uncontrollable gradient optimization process. 
In contrast, we view the semantic dependencies within complex prompts as a neural network, enabling the application of advanced deep learning methods. 
(1) Unlike traditional ML approaches, our method employs Textual Dropout, which randomly discards certain prompt sentences to improve robustness. 
Additionally, we use Textual Learning Rate and Textual Simulated Annealing to better control the optimization direction, enhancing robustness over monotonic gradient optimization.
(2) In deep learning, optimization speed often depends on the gradient optimization difficulty. To accelerate convergence, we apply Textual Learning Rate Decay, Textual Momentum, and Textual Contrastive Learning for more efficient gradient optimization.
(3) Unlike the single prompt modularization in the ML paradigm, our method integrates Textual Regularization to prevent overcomplication of prompt sentences, ensuring that the model remains concise. The correspondence between all seven methods we propose and classical deep learning methods is shown in Table \ref{tab:dl-dltg-comparison}.

To evaluate the effectiveness of our proposed approach, we conduct extensive experiments on five datasets covering multiple tasks. The results indicate that our method significantly enhances the stability, efficiency, and generalizability of prompt optimization. Notably, it outperforms the previous state-of-the-art method by 8.1\% and even surpasses manually designed prompts, demonstrating its superiority in automated prompt optimization.

The contributions of this work are summarized as follows:
\begin{itemize}[leftmargin=2ex,itemsep=0ex]
    \item We first identify key challenges in robustness, efficiency, and generalization in LLM-based prompt optimization, revealing the limitations of existing methods through both theoretical and empirical analysis.
    \item Drawing inspiration from traditional deep learning techniques, we propose \ours to enhance robustness, efficiency, and generalization by incorporating 7 novel text-based gradient optimization strategies.
    \item We conduct extensive experiments across diverse datasets to validate our approach, providing a comprehensive understanding of prompt optimization challenges and best practices.
\end{itemize}

\begin{figure}[t!]
    \centering
    \includegraphics[width=1\linewidth]{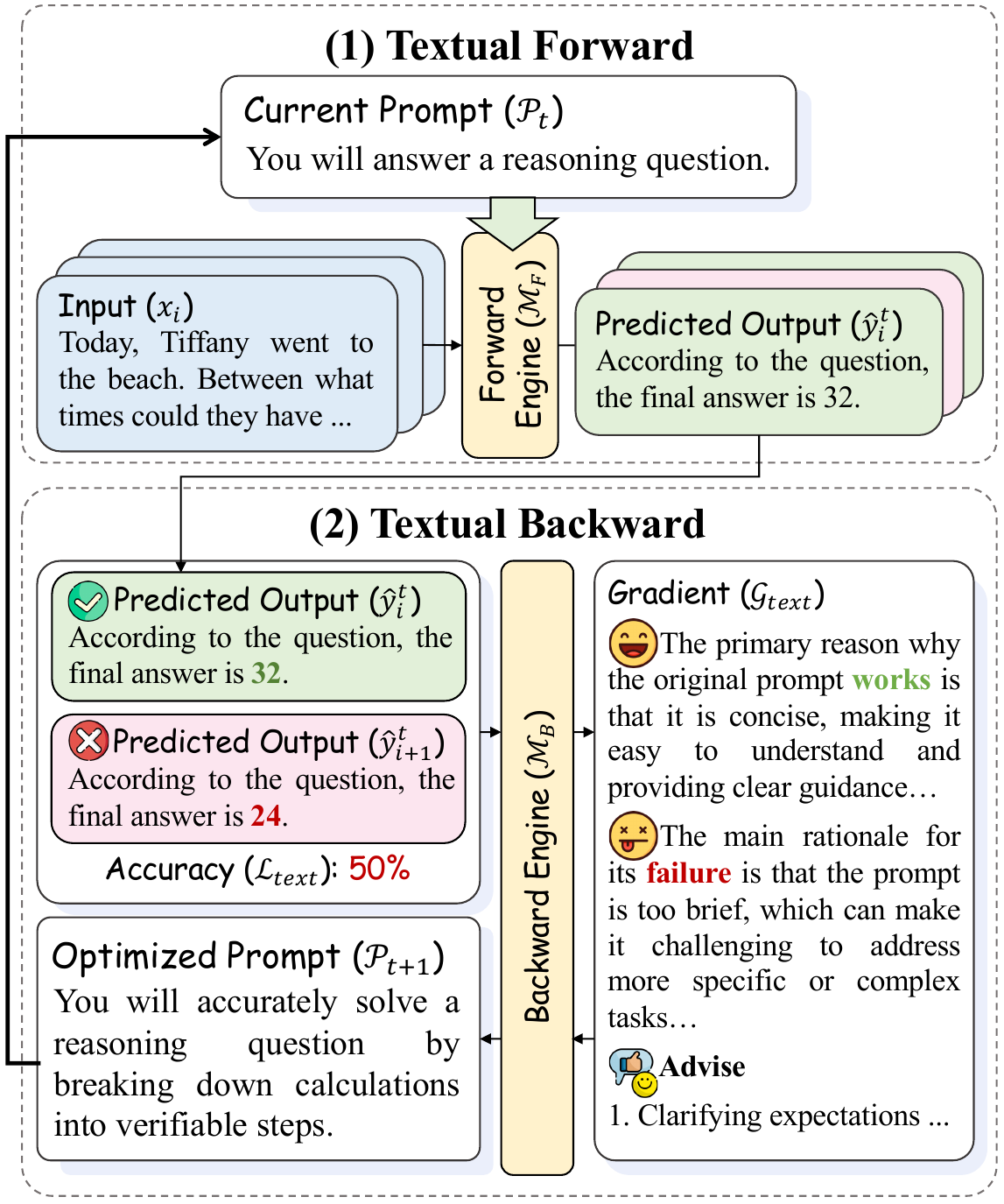}
    %\hspace{-0.05\linewidth}
    \caption{Current reflection-based paradigm for prompt optimization.\vspace{-6pt}}
    \label{fig:reflection-based-paradigm}
\end{figure}
\begin{table*}[t]
\scriptsize
\centering
\renewcommand{\arraystretch}{1.2}
\begin{adjustbox}{width=\textwidth}
\begin{tabular}{p{2.07cm}lcc}
\toprule
Traditional Methods & \ours Methods & Description \\
\midrule
Learning Rate & Textual Learning Rate (\textsc{Tlr}) & \multicolumn{1}{m{8cm}}{ \textsc{Tlr} controls the number of sentence modifications per step to stabilize updates.} \\

Dropout  & Textual Dropout (\textsc{Tdo}) & \multicolumn{1}{m{8cm}}{  \textsc{Tdo} randomly skips sentence updates to reduce overfitting and preserve beneficial modifications.} \\

Simulated Annealing & Textual Simulated Annealing (\textsc{Tsa}) & \multicolumn{1}{m{8cm}}{ \textsc{Tsa} uses training accuracy as energy, accepting suboptimal solutions probabilistically. }\\

Learning Rate Decay & Textual Learning Rate Decay (\textsc{Tlrd}) & \multicolumn{1}{m{8cm}}{ \textsc{Tlrd} enables early-stage exploration followed by gradual refinement of prompts. }\\

Momentum & Textual Momentum (\textsc{Tm}nt) & \multicolumn{1}{m{8cm}}{ \textsc{Tm}nt utilizes past gradients to smooth updates and enhance efficiency. }\\

Contrastive Learning & Textual Contrastive Learning (\textsc{Tcl}) & \multicolumn{1}{m{8cm}}{ \textsc{Tcl} differentiates high- and low-quality prompts to encourage effective patterns. }\\

L1/L2 Regularization & Textual Regularization (\textsc{Tr}egu) & \multicolumn{1}{m{8cm}}{ \textsc{Tr}egu removes redundant phrases and simplifies sentence structures for better generalization. }\\
\bottomrule
\end{tabular}
\end{adjustbox}
\caption{A brief comparison of traditional deep learning methods and their textual counterparts in DLPO. Some methods adapt numerical optimization concepts to the text space through sentence-level operationalization.}
\label{tab:dl-dltg-comparison}
\end{table*}

\section{Problem Formalization}
\subsection{Prompt Optimization}
As shown in Figure~\ref{fig:reflection-based-paradigm}, the paradigm of prompt optimization leverages LLMs as optimizers by refining and enhancing prompts based on feedback from external environments. Formally, we consider a training dataset $\mathcal{D} = {(x_i, y_i)}_{i=1}^n$ drawn from a joint distribution $\mathbb{D}(X, Y)$.

\paragraph{Textual Forward}
Given an input-output pair $(x_i, y_i)$ and current prompt $\mathcal{P}_t$, the LLM $\mathcal{M}_F$, acts as forward engine, generating a predicted output:\vspace{-5pt}
\begin{equation}
  \hat{y}^t_i = \mathcal{M}_F(y_i|x_i, \mathcal{P}_t).
\end{equation}

\paragraph{Textual Backward}
To quantify the discrepancy between the predicted output $\hat{y}^t_i$ and the ground truth $y_i$, we define the textual loss function as:\vspace{-5pt}
\begin{equation} 
  \mathcal{L}_{text} = \mathcal{F}_{loss}(\{\hat{y}^t_1, y_1\},\dots, \{\hat{y}^t_i, y_i\}, \dots),
\end{equation}
where $\mathcal{F}_{loss}(\cdot)$ denotes the human-defined loss function, such as accuracy or model-generated score. 

To guide the improvement of the prompt $\mathcal{P}_t$, textual gradient is defined to determine the direction of optimization.
Here, the LLM acts as an implicit gradient estimator, computing the gradient through reflection: \vspace{-5pt}
\begin{equation} 
  \mathcal{G}_{text} = \frac{\partial \mathcal{L}_{text}}{\partial \mathcal{P}_t} = \mathcal{M}_{B}(\mathcal{G}|x_i, y_i, \mathcal{P}_t, \mathcal{L}_{text}),
\end{equation}
where $\mathcal{M}_{B}$ denotes the LLM that functions as a backward engine. It is always considered more powerful compared to the forward engine.

After textual gradient computation, the prompt $\mathcal{P}_t$ is iteratively updated using the textual gradient to minimize the loss. At this stage, the LLM functions as the optimizer, generating an improved prompt $\hat{\mathcal{P}}_{t+1}$ as: \vspace{-5pt}
\begin{equation} 
  \hat{\mathcal{P}}_{t+1} = \mathcal{M}_{B}(\mathcal{P}_{t+1}|x_i, y_i, \mathcal{P}_t, \mathcal{G}_{text}). 
\end{equation}

\subsection{Prompt Application}

Once optimized, the prompt is applied to previously unseen test data $x'$ to generate the predicted output: \vspace{-5pt}
\begin{equation} 
  \hat{y}'= \mathcal{M}(y'|x', \mathcal{P}^*).
\end{equation}
where $\mathcal{P}^*$ refers to the optimized prompt.
Our primary framework for implementing prompt optimization is TextGrad~\citep{yuksekgonul2024textgrad}, a widely recognized prompt optimization framework. 
It modularly implements the entire pipeline outlined in this section, making it particularly well-suited for our analysis.
\section{Experiment Setups}
We utilize several benchmarks to evaluate the effectiveness of our proposed methods,
including GSM8K~\citep{cobbe2021gsm8k}, MATH~\citep{hendrycksmath2021}, BigGSM (BGSM)~\citep{chen2024unlocking}, BigBenchHard Object Counting (BBH)~\citep{suzgun2022challenging}, and MGSM~\citep{shi2022language}. 
The detailed descriptions can be found in Appendix~\ref{appendix:benchmark}. 

\begin{figure*}[t]
    \centering
    \hspace{-0.03\linewidth}
    \subfloat[\texttt{BBH}]{%
        \includegraphics[width=0.261\linewidth]{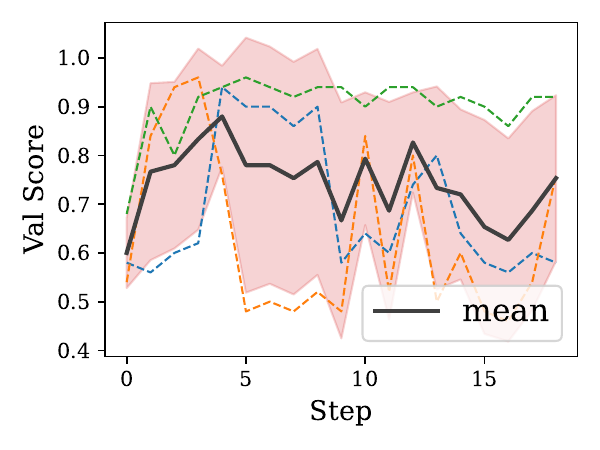}
    }
    \hspace{-0.021\linewidth}
    % \hfill
    \subfloat[\texttt{GSM8k}]{%
        \includegraphics[width=0.261\linewidth]{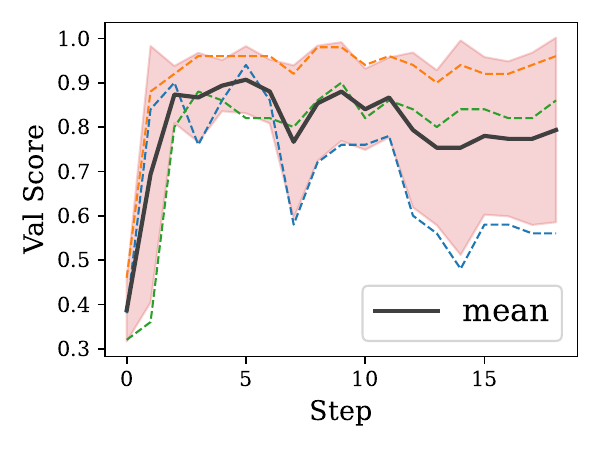}
    }
    \hspace{-0.021\linewidth}
    % \hfill
    \subfloat[\texttt{BigGSM}]{%
        \includegraphics[width=0.261\linewidth]{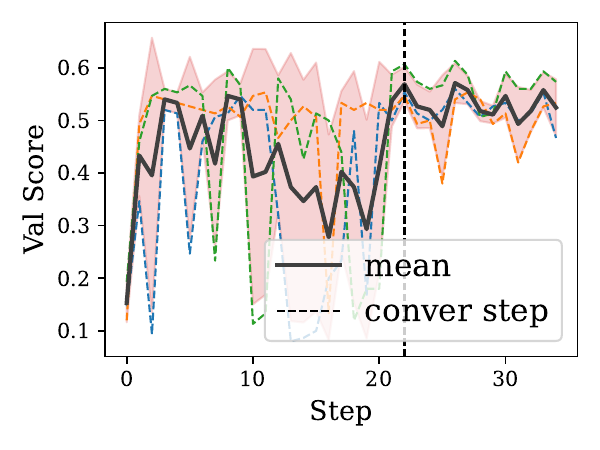}
    }
    \hspace{-0.021\linewidth}
    % \hfill
    \subfloat[\texttt{GSM8k-Val-Train}]{%
        \includegraphics[width=0.261\linewidth]{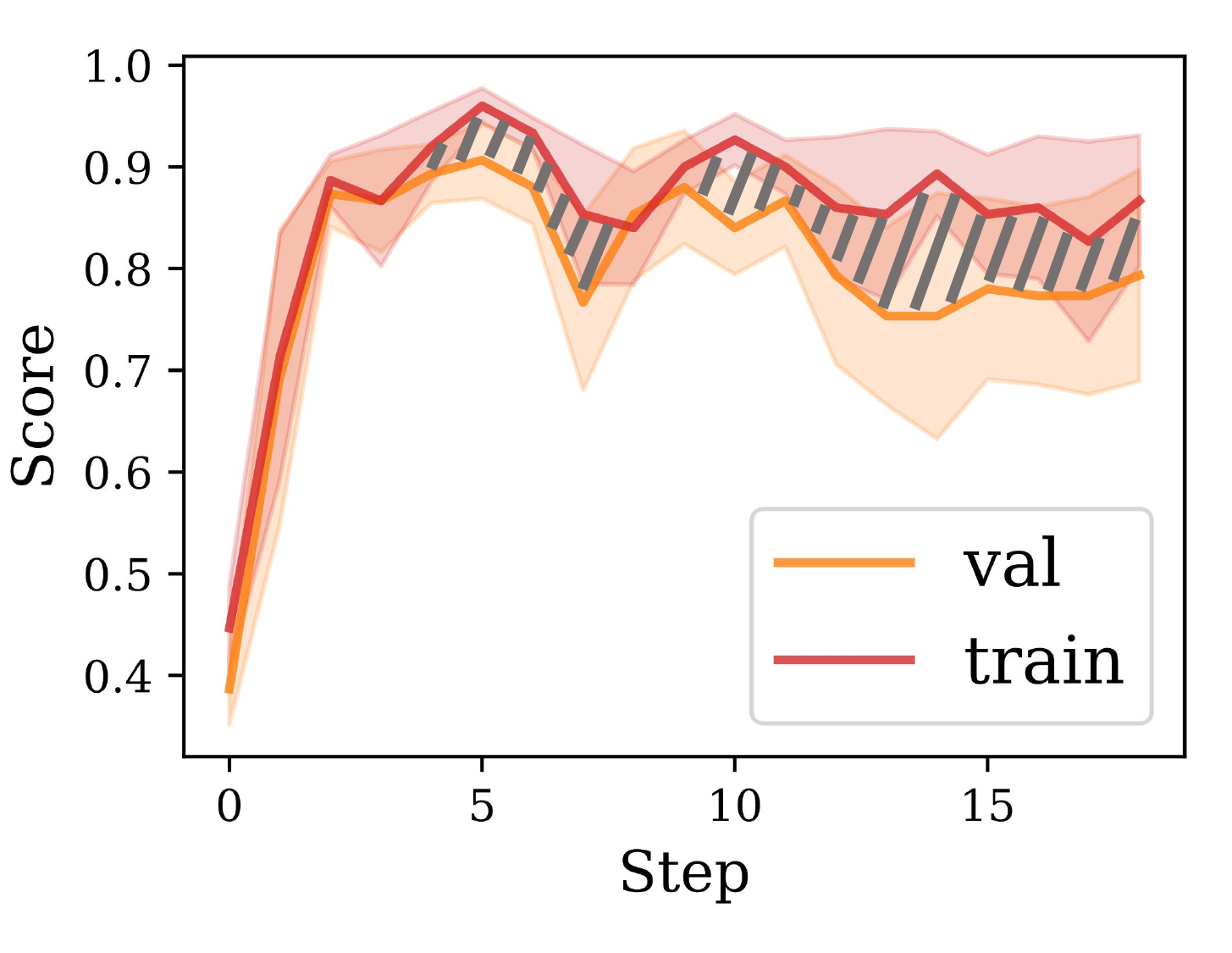}
    }
    
    \caption{\textbf{a}, \textbf{b}, \textbf{c} respectively show the validation-set accuracy of 3 different seeds and their mean values and standard deviations on BBH, GSM8K, and BigGSM environments. The mean values are represented by black solid lines, and the standard deviations are indicated by red shaded areas. \textbf{d} shows the \textbf{\textcolor{red}{Training-set}} and \textbf{\textcolor{orange}{Validation-set}} mean accuracy results of 3 different seeds on GSM8K, along with their standard deviations. To make the image clearer, we use $\frac{1}{2}$ standard deviation as the shaded area.}
    \label{fig:empirical_analysis}
  \end{figure*}
\section{Preliminary Analysis}

In this section, we explore the following three key drawbacks for current reflection based prompt optimization paradigm.

\subsection{Lack of Robustness}
The robustness of an optimization process is a critical prerequisite for both convergence and generalization. In this context, we define robustness as the stability of the optimization process. As illustrated in Fig.~\ref{fig:empirical_analysis} (a,b), the results reveal significant instability. Specifically, variations in random seeds lead to substantial discrepancies in convergence, with the variance at the final step reaching 20.8\% for the GSM8K environment. Moreover, in the BBH, the current paradigm fails to achieve convergence even by the end of the training process.

\subsection{Low Efficiency}
Efficiency is defined here as the convergence speed during prompt optimization. As illustrated in Fig.~\ref{fig:empirical_analysis} (c), we aim for the prompt optimizer to achieve rapid convergence to a stable value. In contrast, the current approach necessitates over 20 iterations to attain its peak performance and reach stabilization. This stark difference underscores the significant inefficiency of the existing method in comparison.

\subsection{Limited Generalizability}

We define \textit{generalizability} as the degree to which LLMs can generalize to test sets or diverse data distributions within the same task domain. As shown in Fig.~\ref{fig:empirical_analysis} (d), the grey shadow reveals a large discrepancy between the validation and training set scores (approximately 10\% accuracy at the optimal point), suggesting that the current update method lacks sufficient generalization capability.

\section{Exploration}
In this section, we solve the above three problems using methods inspired by traditional Deep Learning step by step.
The detailed implementation of our methods can be found in appendix \ref{appen:details}.

\subsection{Exploration for Robustness}
\label{sec:exploration_for_robustness}
The robustness of optimization is one of the most critical issues in subsequent work. 
Only by ensuring the robustness of optimization can other problems and methods be deemed worthy of consideration and meaningful.

We meticulously analyze the optimized sequence of prompts and observe that the variations between updates are highly significant. 
After just a single update, the prompts appear entirely different compared to their previous state.
To address this issue, we propose three effective methods: Textual Learning Rate (\textsc{Tlr}) , Textual Dropout (\textsc{Tdo}) and Textual Simulated Annealing (\textsc{Tsa}).

\begin{figure}[t]
    \centering
    \hspace{-0.05\linewidth}
    \subfloat[\texttt{GSM8K-\textsc{Tlr}}]{
        \includegraphics[width=0.51\linewidth]{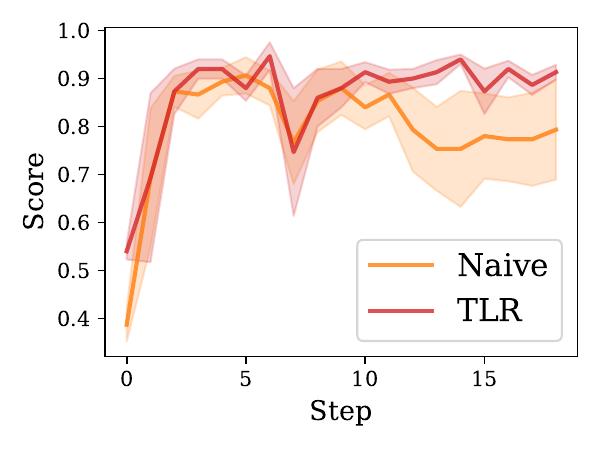}
    }
    \hspace{-0.05\linewidth}
    \subfloat[\texttt{BBH-\textsc{Tsa}-\textsc{Tlr}}]{
        \includegraphics[width=0.51\linewidth]{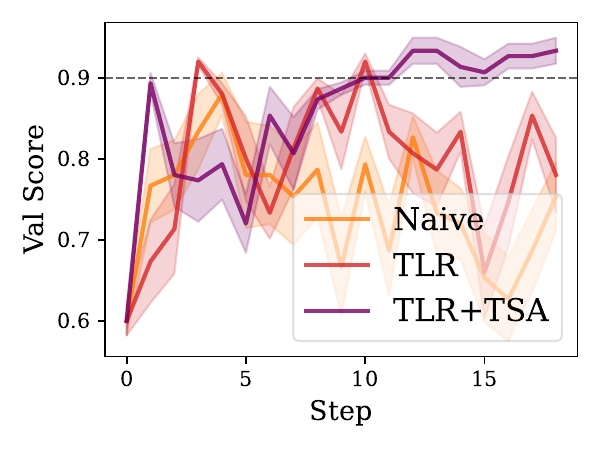}
    }
    \caption{\textbf{a}, mean results of 3 different seeds for \textcolor{red}{\textsc{Tlr}+\textsc{Tdo}} and \textcolor{orange}{Naive} on GSM8K. We use $\frac{1}{2}$ of the standard deviation as the shaded area. \textbf{b}, mean results of 3 different seeds for \textcolor{myPurple}{\textsc{Tsa}+\textsc{Tlr}+\textsc{Tdo}}, \textcolor{red}{\textsc{Tlr}+\textsc{Tdo}} and \textcolor{orange}{Naive} on BBH. To make the image clearer, we use $\frac{1}{4}$ of the standard deviation as the shaded area.}
    \label{fig:robustness_comparison}
\end{figure}

\subsubsection{Textual Learning Rate \& Dropout}
\paragraph{Textual Learning Rate (\textsc{Tlr})}
Traditional ML-based PO often disregards the gradient's magnitude, impeding convergence to the optimal solution.
To address this, drawing inspiration from DL, we propose a Textual Learning Rate to regulate the extent of gradient updates.
Specifically, each modification, deletion, or addition of a sentence in a prompt constitutes an update unit.
The \textsc{Tlr} value, denoted as \( \mathcal{R} \), sets an upper limit on the number of update units per step.
If an LLM attempts to exceed \( \mathcal{R} \), we prompt LLMs to prioritize impactful changes while discarding non-essential ones.

\paragraph{Textual Dropout (\textsc{Tdo})}
Traditional ML-based PO treats the prompt as an indivisible whole, making local optimization challenging and leading to large updates that compromise robustness. To address this, we adopt a DL perspective, viewing the prompt as a structured semantic network. Specifically, we introduce Textual Dropout (\textsc{Tdo}), which selectively drops or skips sentence updates to prevent full parameter optimization.

Formally, given a prompt $\mathcal{P}_t$ with \( \mathcal{S} \) sentences and a dropout rate \( p \) , the number of preserved sentences \( \mathcal{K} \) is:
\begin{equation}
  \mathcal{K} = \lceil p \cdot \mathcal{S} \rceil.
\end{equation}
The updated prompt \(\hat{\mathcal{P}}_{t+1}\) is:  \vspace{-4pt}
\[
    \hat{\mathcal{P}}_{t+1} = \mathcal{P}_{\text{preserved}} \oplus  \mathcal{P}_{\text{updated}},
\]
where $\oplus$ denotes merge operation. Here, \(\mathcal{P}_{\text{preserved}}\) consists of the $\mathcal{K}$ preserved sentences, while \(\mathcal{P}_{\text{updated}}\) includes the modified sentences based on the textual gradient \(\mathcal{G}_{\text{text}}\).

Since both \textsc{Tlr} and \textsc{Tdo} regulate gradient updates at the sentence level and encourage a certain degree of restriction on the magnitude of single-step updates, we apply them jointly to evaluate their effectiveness.
As shown in Fig.~\ref{fig:robustness_comparison} (a,b), experiments on GSM8K and BBH demonstrate that \textsc{Tlr}+\textsc{Tdo} outperforms the naive approach and significantly enhances optimization robustness.

\subsubsection{Textual Simulated Annealing \textsc{Tsa}}

To further enhance the gradient optimization in DL views, we propose Textual Simulated Annealing scheme to guarantee stable update during training by dynamically adjusting prompts based on their accuracy to effectively control the optimization direction.
Simultaneously, to prevent the model from getting stuck in local optima, \textsc{Tsa} encourages the model to accept suboptimal solutions with a certain probability.
Specifically, we define training set accuracy as the ``Energy'' function ($\mathcal{E}(\cdot)$), and compare the accuracy before and after each update. 
If accuracy decreases, the suboptimal solution is accepted with a probability determined by:  \vspace{-4pt}
\begin{equation}
  P(\Delta \mathcal{E}, \mathcal{T}) = \exp\left(\frac{\Delta \mathcal{E}}{\mathcal{T}}\right)
\end{equation}
where\( \Delta \mathcal{E} = \mathcal{E}(x') - \mathcal{E}(x) \) is the accuracy change between the new solution \( x' \), derived from optimized prompt $\hat{\mathcal{P}}_{t+1}$, and the current solution \( x \), generated with original prompt $\mathcal{P}_{t}$. Here, the temperature \( \mathcal{T} \) will gradually decrease with each update.
As \( \mathcal{T} \) reduces, the probability of accepting suboptimal solutions decreases, enabling the model to escape local optima in the early stages and converge to better solutions later. Building on this, \textsc{Tsa} ensures stable and robust accuracy convergence.

In Fig.~\ref{fig:robustness_comparison} (b), it is shown that while \textsc{Tlr} methods outperform the Naive approach in BBH, they still exhibit substantial variability. To address this issue, we integrate \textsc{Tsa} into the \textsc{Tlr} approach. The results indicate that the incorporation of \textsc{Tsa} into \textsc{Tlr} significantly improves both performances and robustness, surpassing the two original baselines.

\begin{takeaways}
\textbf{Takeaway:}
\textit{Limiting the magnitude, updated parameter size, and direction of each update step to some extent can enhance the robustness and stability of the PO.}

\end{takeaways}

\subsection{Exploration for Efficiency}
\label{sec:exploration_for_efficiency}

Through our definitions of \textsc{Tlr}, \textsc{Tdo}, and \textsc{Tsa}, we have largely mitigated the issue of update stability.
Next, we shift our focus to convergence speed.
To enhance convergence efficiency, we propose three methods: Textual Learning Rate Decay (\textsc{Tlrd}), Textual Momentum (\textsc{Tm}nt) and Textual Contrastive Learning (\textsc{Tcl})

\begin{figure}[t]
    \centering
    \hspace{-0.05\linewidth}
    \subfloat[\texttt{BigGSM-\textsc{Tlrd}}]{
        \includegraphics[width=0.51\linewidth]{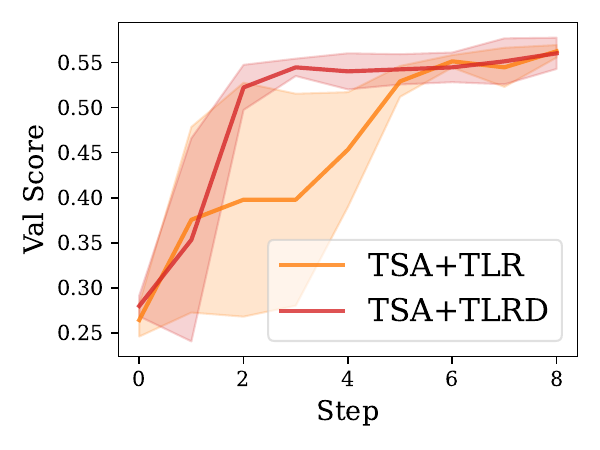}
    }
    \hspace{-0.05\linewidth}
    \subfloat[\texttt{MGSM-\textsc{Tlrd}}]{
        \includegraphics[width=0.51\linewidth]{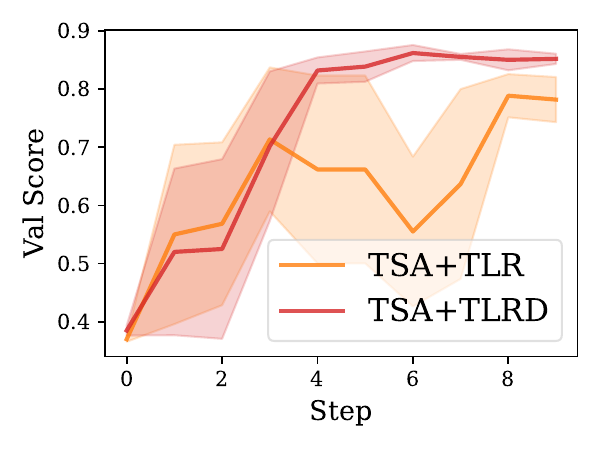}
    }
    \caption{\textbf{a}, \textbf{b}, mean results of 3 different seeds for \textcolor{red}{\textsc{Tsa}+\textsc{Tlrd}} and \textcolor{orange}{\textsc{Tsa}+\textsc{Tlr}} on validation set of BigGSM and MGSM environment. We use $\frac{1}{2}$ of the standard deviation as the shaded area.}
    \label{fig:efficiency_comparison}
\end{figure}
\subsubsection{Textual Learning Rate Decay (\textsc{Tlrd})}

Inspired by the role of learning rate decay in deep learning, where larger updates facilitate exploration in early stages and smaller updates ensure fine-tuning later, we define \textsc{Tlrd} to enhance convergence speed and stability in prompt optimization. At the beginning of the update process, a higher learning rate encourages substantial modifications to the initial prompt, enabling diverse exploration of potential solutions. As the optimization progresses and the prompt approaches a satisfactory level, the learning rate decreases dynamically to focus on fine-grained refinements. 

As shown in Fig.~\ref{fig:efficiency_comparison} (a,b), we compare the fixed T\textsc{lr} (\( \mathcal{R} = 1\)) with the learning rate decay strategy in the BigGSM and MGSM.
The results demonstrate that the learning rate decay method achieves faster convergence compared to the fixed learning rate approach.

\subsubsection{Textual Momentum (\textsc{Tm}nt)}
During each update, the model samples batches independently, without incorporating information from previous ones. This can cause notable bias in gradient optimization, particularly when the batch size is small, as batch-to-batch variations can significantly impact the convergence trajectory.

To mitigate this issue, we draw inspiration from classical machine learning techniques and introduce Textual Momentum (\textsc{Tm}nt).
By incorporating feedback from past updates, our approach allows the optimizer to recognize previous gradient directions and integrate them into the current update, ultimately refining the final gradient direction for more stable optimization.
The final ideal gradient direction \( \mathcal{G}_{\text{final}} \) is computed as:  \vspace{-4pt}
\begin{equation}
  \mathcal{G}_{\text{final}} = \mathcal{G}_{\text{current}} + \sum_{i=1}^{3} \gamma^{i} \cdot \mathcal{G}_{\text{past}_i}
\end{equation}
where \( \mathcal{G}_{\text{current}} \) is the current batch gradient, \( \mathcal{G}_{\text{past}_i} \) represents the \( i \)-th previous gradient, and \( \gamma \) is a decay factor (e.g., \( \gamma = 0.9 \)) reducing older gradients' influence. In practice, we do not explicitly define certain details, such as the decay coefficient \( \gamma \). Instead, we only present the last three feedbacks and encourage the optimizer to focus more on updates closer to the current step to balance recent trends without over-relying on outdated information.

\begin{figure}[t]
    \centering
    \hspace{-0.05\linewidth}
    \subfloat[\texttt{BigGSM}]{
        \includegraphics[width=0.51\linewidth]{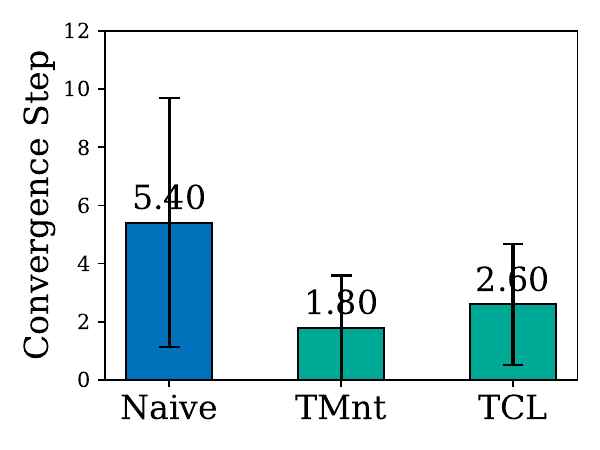}
    }
    \hspace{-0.05\linewidth}
    \subfloat[\texttt{MGSM}]{
        \includegraphics[width=0.51\linewidth]{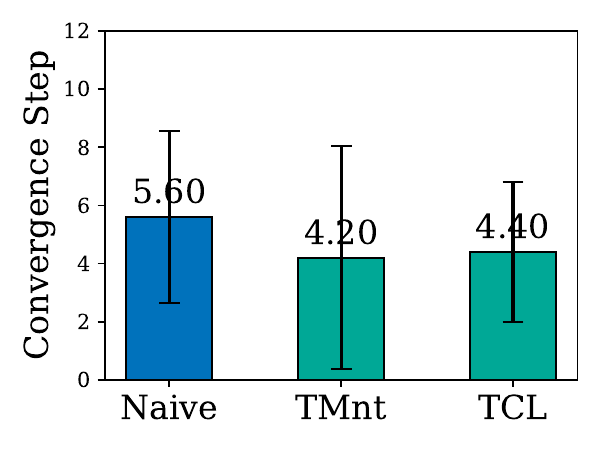}
    }
    \caption{We tested the \textsc{Tm}nt and \textsc{Tcl} methods on the validation set of BigGSM and MGSM environment. The convergence steps and their standard deviations are displayed in the bar chart.}
    \label{fig:mnt_cl_comparison}
\end{figure}

As shown in Fig.~\ref{fig:mnt_cl_comparison} (a,b), based on our previously proposed \textsc{Tlr} method, we compare the pure \textsc{Tlr} with \textsc{Tlr}+\textsc{Tm}nt on the validation sets of BigGSM and MGSM. 
It is evident that after adding the \textsc{Tm}nt method, the convergence speed has improved.

\subsubsection{Textual Contrastive Learning (\textsc{Tcl})}

Building on the success of contrastive learning in improving model efficiency in DL view~\citep{pmlr-v139-radford21a,qin-etal-2022-gl}, we propose Textual Contrastive Learning (\textsc{Tcl}), which encourages the model to differentiate high-quality (\textit{positive}) prompts from suboptimal (\textit{negative}) ones, thus enhancing the optimization process.

Specifically, we first evaluate each updated prompt on the training set to obtain its accuracy as a performance metric.  
Let \( \mathcal{A}(x) \) denote the accuracy of prompt \( x \) on the training set.  
Next, we categorize historical prompts into \textit{positive} (higher accuracy) and \textit{negative} (lower accuracy) groups.  
We then sample one prompt from each group based on an accuracy-weighted probability distribution.  
If the accuracy difference between the positive and negative prompts falls below a predefined margin, we only encourage the LLM to learn from the positive prompt.  
Otherwise, we guide the optimizer to: (1) Imitate distinguishing features of positive prompts, represented by the set \( \mathcal{F}_{+} \). (2) Retain shared features between positive and negative prompts, represented by \( \mathcal{F}_{\cap} \). (3) Eliminate undesirable features unique to negative prompts, represented by \( \mathcal{F}_{-} \).

The final gradient update \( \mathcal{G}_{\text{final}} \) is computed as:  \vspace{-4pt}
\begin{equation}
\mathcal{G}_{\text{final}} = \mathcal{G}_{\text{current}} + \mathcal{F}_{+} - \mathcal{F}_{-}
\end{equation}

By refining the gradient in this manner, the optimizer learns to reinforce beneficial prompt features while suppressing undesirable ones, ultimately improving the quality of future updates.

To test the effectiveness of \textsc{Tcl}, we compare the convergence step of Naive and \textsc{Tcl} methods on the validation sets of BigGSM and MGSM.
As shown in Figure \ref{fig:mnt_cl_comparison} (a,b), it is evident that after adding the \textsc{Tcl} method to naive, the convergence speed have consistently improved. 
Unfortunately, we also observe that although \textsc{Tm}nt appears to outperform the \textsc{Tcl} method in the figure, with a faster convergence speed, its final convergence effect is inferior to that of the \textsc{Tcl} method and exhibits higher instability.
\begin{takeaways}
    \textbf{Takeaway:}
    \textit{Incorporating dynamic, historical, and contrasting directions of gradient optimization from previous update steps can significantly accelerate the convergence speed.}

    \end{takeaways}
\subsection{Exploration for Generalizablity}

\label{sec:exploration_for_generalizablity}
Before addressing generalizability, we first propose several methods to enhance model update stability and accelerate convergence.
However, our ultimate goal is to improve the model's overall performance, particularly its ability to generalize beyond the training set.
Specifically, we aim to extend the learned results to:
(1) The validation and test sets of the same dataset.
(2) Datasets with similar content but varying levels of difficulty.
(3) Datasets involving similar tasks.

Therefore, we train on a subset of the GSM8K training set to derive the final prompt and evaluate its performance on the test sets of GSM8K, BigGSM, BBH, and MATH.
Building on the previously proposed effective methods, we adjust the training set size and batch size to examine their impact on the generalizability of our approach.
Additionally, we integrate prior techniques and introduce Textual regularization to further explore its role in enhancing generalizability.
More experiments on generalization to different LLM models (Gemini \citep{team2023gemini}, DeepSeek-R1 \citep{guo2025deepseek}) are presented in the appendix \ref{appendix:gentodif_model}.
\begin{table}[t]
    \centering
    \fontsize{8pt}{9pt}\selectfont
    \begin{tabular}{l|l|cccc}
    \toprule
    expname & hyper &  GSM8K & BGSM & BBH & MATH \\
    \midrule
    batch size    & bs = 3    & 93.3  & \textbf{57.3}  & 79.7 & \textbf{73.9 }\\
    & bs = 6    & 93.2  & 52.7  & \textbf{85.0} & \textcolor{red}{65.6} \\
    & bs = 9    & 93.3  & 57.2  & \textcolor{red}{77.0}  & 70.2 \\
    & bs = 12  & \textbf{93.7}  & 54.8  & 79.7 & 69.6 \\
    & bs = 15  & \textcolor{red}{89.7}  & \textcolor{red}{48.4} & 80.0 & 70.8 \\
\midrule
    training set & ts = 50    & 93.2  & 52.7  & 85.0 & 65.6 \\
    size & ts = 100 & \textbf{94.3}  & \textbf{57.0}  & \textbf{90.3}  & \textbf{69.7} \\
    \bottomrule
    \end{tabular}
    \caption{Accuracy comparison across different batch sizes and training set sizes. All results are the average of three different seeds. The \textbf{bold} is the best while the \textcolor{red}{red} is the worst. More details are shown in the Table \ref{tab:bsts_std}.}
    \label{tab:bsts}
\end{table}
\subsubsection{Exploration for Training Set Size}
Inspired by the training scaling experience in DL~\cite{kaplan2020scaling}, we manage to scale up the training set size to improve the generalization.
As shown in Table~\ref{tab:bsts}, we create two training subsets by sampling 50 and 100 instances from GSM8K. The model trained on 100 instances consistently outperforms the one trained on 50 across all test sets. This aligns with intuition, as a larger training set offers broader sample coverage, reducing distribution shift and improving generalizability.

\subsubsection{Exploration for Batch Size}

To explore the impact of different batch sizes on generalizability, we evaluate five batch sizes in terms of final test accuracy, as shown in Table~\ref{tab:bsts}. The results indicate that:
(1) Within a certain range, a smaller batch size can achieve better performance, but increasing the batch size has a negligible effect on performance.
(2) However, if the batch size is excessively large, resulting in too few update steps, the optimization process may fail to converge to the optimal solution.

It is worth noting that the total number of loss calculations remains constant across different batch sizes, while the number of backward passes and updates varies. Therefore, when the requirements for final performance are not overly stringent, it is advisable to appropriately increase the batch size. This reduces the frequency of backward passes and updates, thereby saving computational resources, such as token usage, without significantly compromising model performance.

\subsubsection{Textual Regularization (\textsc{Tr}egu)}
\begin{table}[t]
    \centering
    \footnotesize 
    \begin{tabular}{l|cccc}
    \toprule
    expname & GSM8K & BGSM & BBH & MATH \\
    \midrule
    w/o \textsc{Tr}egu  &93.3 &55.0 &81.0& 67.9 \\
    w/ \textsc{Tr}egu  &\textbf{94.0} &\textbf{55.9} &\textbf{87.2} &\textbf{70.5} \\
    \bottomrule
    \end{tabular}
    \caption{We compare the performance with and without the addition of \textsc{Tr}egu on the baseline methods of \textsc{Tsa}, \textsc{Tlrd}, and \textsc{Tcl}. The final results are obtained by averaging the outcomes across the six seeds. The \textbf{Bold} data indicates the highest value in each column.}
    \label{tab:regu}
\end{table}

Ockham's razor in DL suggests that simpler models with fewer parameters often generalize better. Building on this principle, we investigate Textual Regularization (\textsc{Tr}egu) techniques to reduce prompt complexity and enhance generalization.
Specifically, textual L2 regularization is applied to reduce prompt complexity, while L1 regularization is used to promote sparsity by eliminating irrelevant features.
Practically, textual L2 regularization encourages simplifying individual sentences, whereas L1 regularization helps remove irrelevant ones.
As shown in Table~\ref{tab:regu}, \textsc{Tr}egu methods demonstrate significant improvements in generalization on various tasks.
\begin{takeaways}
    \textbf{Takeaway:}
    \textit{A larger training dataset and a more precise and streamlined language structure contribute to enhancing the generalizability of the resulting prompt.}

\end{takeaways}

\section{Best Practices}

By systematically integrating various optimization methods, we progressively improve update robustness, generalization, and efficiency.
However, identifying the optimal combination of these techniques remains an open challenge.
To address this, we conduct extensive experiments on BigGSM, MGSM, and BBH, evaluating different method combinations to determine the most effective approach.
Our findings indicate that the optimal configuration includes \textsc{Tlrd}, \textsc{Tsa}, \textsc{Tcl}, and \textsc{Tregu}, collectively referred to as \texttt{DLPO}.
To assess its effectiveness, we compared our approach against several baselines:
(1) TextGrad (TG)~\citep{yuksekgonul2024textgrad}, a gradient-based prompt optimization method.
(2) APO~\citep{pryzant2023automatic}, a widely recognized automatic prompt optimization technique.
(3) HUMAN, a human-optimized prompt method derived from the BigGSM dataset.

As shown in Table~\ref{tab:bestprac}, our method demonstrates state-of-the-art performance on the test sets of both the training datasets and external datasets. 
It consistently outperforms other methods, including TG and APO, which exhibit lower accuracy across most datasets. 
This indicates that our method introduces more effective optimization strategies for enhancing prompt quality. 
Even when compared to human-optimized prompts, our approach still achieves superior results.

\begin{table}[t]
    \centering
    \footnotesize 
    \begin{tabular}{c|cccc}
    \midrule
    \rowcolor{gray!8}\multicolumn{5}{c}{Trainset : BigGSM} \\
    \midrule
    Method & BGSM & BBH & MATH & GSM8K \\
    \midrule

    TG & 55.7 & 87.3 & 69.7 & 92.7 \\

    APO & \underline{58.4} & \underline{89.0} & 59.7 & \textbf{93.4} \\
    DLPO & \textbf{60.2} & \textbf{89.7} & \underline{71.3} & \underline{93.3} \\
    \midrule
    HUMAN & 54.4 & 89.0  & \textbf{72.0} & 90.3 \\
    \midrule
    \rowcolor{gray!8}\multicolumn{5}{c}{Trainset : MGSM} \\
    \midrule
    Method & MGSM & BBH & MATH & BGSM \\
    \midrule

    TG & 61.8 & 77.7 & \underline{68.8} & 32.5 \\

    APO & \underline{82.7} & \underline{86.0} & 51.0 & \underline{47.9} \\
    DLPO & \textbf{86.7} & \textbf{89.3} & \textbf{70.8} & \textbf{55.6} \\
    \midrule
    \rowcolor{gray!8}\multicolumn{5}{c}{Trainset : BBH object counting} \\
    \midrule
    Method & BBH & GSM8K & MATH & BGSM \\
    \midrule

    TG & 63.8 & 83.3 & \textbf{72.2} & 44.8 \\

    APO & \underline{90.9} & \underline{92.8} & 68.9 & \underline{55.8} \\
    DLPO & \textbf{94.2} & \textbf{93.9} & \underline{71.9} & \textbf{56.7} \\
    \midrule
    \end{tabular}
    \caption{We test various methods on the trainset of BigGSM, MGSM, and BBH. Except for the HUMAN method, where the prompt is taken directly from the original article, each result represents the average of three different seeds for the corresponding method. The \textbf{bold} data indicates the highest value in each column, while the \underline{underlined} data represents the second-highest value. The complete table is in the appendix \ref{tab:bstp_std}.}

    \label{tab:bestprac}
\end{table}
\section{Related Work}

Early prompt optimization methods relied on internal access to LLMs \citep{lester2021power, li2021prefix, qin2021learning, gu2021ppt, liu2021p}.
However, these methods often require access to the logits or internal states of LLMs, which is infeasible for those only accessible through APIs \citep{hou2023promptboosting, tang2024unleashing}.
To overcome these limitations, recent research has shifted toward exploring gradient-free methods, such as
reinforcement learning based methods \citep{deng2022rlprompt, zhang2022tempera, diao2022black},
search-based methods \citep{prasad2022grips, pryzant2023automatic}
and other techniques like evolutionary algorithms \citep{sun2022black} and boosting \citep{hou2023promptboosting}.
Among them, Reflection-based methods have gained significant attention due to their ability to iteratively refine prompts while maintaining interpretability.
Several notable frameworks exemplify this paradigm, such as
ProTeGi(APO) \citep{pryzant2023automatic}, TextGrad \citep{yuksekgonul2024textgrad}, PromptAgent \citep{wang2023promptagent}, PE2 \citep{ye2023prompt}, and GPO \citep{tang2024unleashing}.
Apart from the above methods, researchers have also explored key challenges in LLM-based prompt optimization. 

\citet{li2023robust} revealed that existing prompt optimization methods are vulnerable to distribution shifts, a critical issue for real-world applications. 
\citet{ma2024large} conducted a comprehensive analysis of LLM-based prompt optimization, introducing a novel method that directly optimizes the target model's behavior in a more controllable manner.

These advances in automated prompt optimization align with our research direction. However, as noted in our problem statement, significant challenges remain in achieving robust, efficient, and generalizable prompt optimization, limiting broader application. We are the first to systematically address these issues, introducing innovative techniques inspired by traditional deep learning to overcome these challenges.

\section{Conclusion}

Through empirical analysis, we identify key limitations in current reflection-based prompt optimization methods, particularly in robustness, efficiency, and generalization. To address these issues, we introduce novel techniques inspired by traditional deep learning paradigms, integrating concepts like gradient modulation, regularization, and adaptive learning into text-based optimization. Our approach enhances stability, convergence speed, and generalizability in a systematic manner. Experiments on diverse tasks and datasets, including BBH, GSM8K, BigGSM, MATH, and MGSM, validate the effectiveness of our methods. By addressing core challenges in prompt optimization, our work provides valuable insights and guides future research in this field.
\section*{Limitations}
Although we have introduced multiple methods and demonstrated their effectiveness, a deeper exploration of the intrinsic properties and potential variants of each method is still lacking. Even for the approaches we have developed, there remains considerable room for enhancement in future research. Additionally, we aspire to see our method applied more broadly, extending beyond prompt optimization to a wider range of fields.

\section*{Acknowledgments}

\bibliography{references}

\begin{thebibliography}{41}
\providecommand{\natexlab}[1]{#1}

\bibitem[{Brown et~al.(2020)Brown, Mann, Ryder, Subbiah, Kaplan, Dhariwal, Neelakantan, Shyam, Sastry, Askell et~al.}]{brown2020language}
Tom Brown, Benjamin Mann, Nick Ryder, Melanie Subbiah, Jared~D Kaplan, Prafulla Dhariwal, Arvind Neelakantan, Pranav Shyam, Girish Sastry, Amanda Askell, et~al. 2020.
\newblock Language models are few-shot learners.
\newblock \emph{Advances in neural information processing systems}, 33:1877--1901.

\bibitem[{Chen et~al.(2024)Chen, Qin, Wang, Zhou, and Che}]{chen2024unlocking}
Qiguang Chen, Libo Qin, Jiaqi Wang, Jinxuan Zhou, and Wanxiang Che. 2024.
\newblock Unlocking the capabilities of thought: A reasoning boundary framework to quantify and optimize chain-of-thought.
\newblock \emph{arXiv preprint arXiv:2410.05695}.

\bibitem[{Cobbe et~al.(2021)Cobbe, Kosaraju, Bavarian, Chen, Jun, Kaiser, Plappert, Tworek, Hilton, Nakano, Hesse, and Schulman}]{cobbe2021gsm8k}
Karl Cobbe, Vineet Kosaraju, Mohammad Bavarian, Mark Chen, Heewoo Jun, Lukasz Kaiser, Matthias Plappert, Jerry Tworek, Jacob Hilton, Reiichiro Nakano, Christopher Hesse, and John Schulman. 2021.
\newblock Training verifiers to solve math word problems.
\newblock \emph{arXiv preprint arXiv:2110.14168}.

\bibitem[{Deng et~al.(2022)Deng, Wang, Hsieh, Wang, Guo, Shu, Song, Xing, and Hu}]{deng2022rlprompt}
Mingkai Deng, Jianyu Wang, Cheng-Ping Hsieh, Yihan Wang, Han Guo, Tianmin Shu, Meng Song, Eric~P Xing, and Zhiting Hu. 2022.
\newblock Rlprompt: Optimizing discrete text prompts with reinforcement learning.
\newblock \emph{arXiv preprint arXiv:2205.12548}.

\bibitem[{Diao et~al.(2022)Diao, Huang, Xu, Li, Lin, Zhou, and Zhang}]{diao2022black}
Shizhe Diao, Zhichao Huang, Ruijia Xu, Xuechun Li, Yong Lin, Xiao Zhou, and Tong Zhang. 2022.
\newblock Black-box prompt learning for pre-trained language models.
\newblock \emph{arXiv preprint arXiv:2201.08531}.

\bibitem[{Du et~al.(2024)Du, Sun, and Snoek}]{du2024ipo}
Yingjun Du, Wenfang Sun, and Cees~GM Snoek. 2024.
\newblock Ipo: Interpretable prompt optimization for vision-language models.
\newblock \emph{arXiv preprint arXiv:2410.15397}.

\bibitem[{Gu et~al.(2021)Gu, Han, Liu, and Huang}]{gu2021ppt}
Yuxian Gu, Xu~Han, Zhiyuan Liu, and Minlie Huang. 2021.
\newblock Ppt: Pre-trained prompt tuning for few-shot learning.
\newblock \emph{arXiv preprint arXiv:2109.04332}.

\bibitem[{Guo et~al.(2025)Guo, Yang, Zhang, Song, Zhang, Xu, Zhu, Ma, Wang, Bi et~al.}]{guo2025deepseek}
Daya Guo, Dejian Yang, Haowei Zhang, Junxiao Song, Ruoyu Zhang, Runxin Xu, Qihao Zhu, Shirong Ma, Peiyi Wang, Xiao Bi, et~al. 2025.
\newblock Deepseek-r1: Incentivizing reasoning capability in llms via reinforcement learning.
\newblock \emph{arXiv preprint arXiv:2501.12948}.

\bibitem[{Hendrycks et~al.(2021)Hendrycks, Burns, Kadavath, Arora, Basart, Tang, Song, and Steinhardt}]{hendrycksmath2021}
Dan Hendrycks, Collin Burns, Saurav Kadavath, Akul Arora, Steven Basart, Eric Tang, Dawn Song, and Jacob Steinhardt. 2021.
\newblock Measuring mathematical problem solving with the math dataset.
\newblock \emph{NeurIPS}.

\bibitem[{Hou et~al.(2023)Hou, O’connor, Andreas, Chang, and Zhang}]{hou2023promptboosting}
Bairu Hou, Joe O’connor, Jacob Andreas, Shiyu Chang, and Yang Zhang. 2023.
\newblock Promptboosting: Black-box text classification with ten forward passes.
\newblock In \emph{International Conference on Machine Learning}, pages 13309--13324. PMLR.

\bibitem[{Kaplan et~al.(2020)Kaplan, McCandlish, Henighan, Brown, Chess, Child, Gray, Radford, Wu, and Amodei}]{kaplan2020scaling}
Jared Kaplan, Sam McCandlish, Tom Henighan, Tom~B Brown, Benjamin Chess, Rewon Child, Scott Gray, Alec Radford, Jeffrey Wu, and Dario Amodei. 2020.
\newblock Scaling laws for neural language models.
\newblock \emph{arXiv preprint arXiv:2001.08361}.

\bibitem[{Lester et~al.(2021)Lester, Al-Rfou, and Constant}]{lester2021power}
Brian Lester, Rami Al-Rfou, and Noah Constant. 2021.
\newblock The power of scale for parameter-efficient prompt tuning.
\newblock \emph{arXiv preprint arXiv:2104.08691}.

\bibitem[{Li et~al.(2023)Li, Wang, Feng, Cao, Zhang, and Chua}]{li2023robust}
Moxin Li, Wenjie Wang, Fuli Feng, Yixin Cao, Jizhi Zhang, and Tat-Seng Chua. 2023.
\newblock Robust prompt optimization for large language models against distribution shifts.
\newblock \emph{arXiv preprint arXiv:2305.13954}.

\bibitem[{Li and Liang(2021)}]{li2021prefix}
Xiang~Lisa Li and Percy Liang. 2021.
\newblock Prefix-tuning: Optimizing continuous prompts for generation.
\newblock \emph{arXiv preprint arXiv:2101.00190}.

\bibitem[{Li et~al.(2024)Li, Peng, He, and Yan}]{li2024evaluating}
Zekun Li, Baolin Peng, Pengcheng He, and Xifeng Yan. 2024.
\newblock Evaluating the instruction-following robustness of large language models to prompt injection.
\newblock In \emph{Proceedings of the 2024 Conference on Empirical Methods in Natural Language Processing}, pages 557--568.

\bibitem[{Liu et~al.(2021)Liu, Ji, Fu, Tam, Du, Yang, and Tang}]{liu2021p}
Xiao Liu, Kaixuan Ji, Yicheng Fu, Weng~Lam Tam, Zhengxiao Du, Zhilin Yang, and Jie Tang. 2021.
\newblock P-tuning v2: Prompt tuning can be comparable to fine-tuning universally across scales and tasks.
\newblock \emph{arXiv preprint arXiv:2110.07602}.

\bibitem[{Ma et~al.(2024)Ma, Wang, Zhou, Li, Du, Gui, Zhang, and Huang}]{ma2024large}
Ruotian Ma, Xiaolei Wang, Xin Zhou, Jian Li, Nan Du, Tao Gui, Qi~Zhang, and Xuanjing Huang. 2024.
\newblock Are large language models good prompt optimizers?
\newblock \emph{arXiv preprint arXiv:2402.02101}.

\bibitem[{Ma{\~n}as et~al.(2024)Ma{\~n}as, Astolfi, Hall, Ross, Urbanek, Williams, Agrawal, Romero-Soriano, and Drozdzal}]{manas2024improving}
Oscar Ma{\~n}as, Pietro Astolfi, Melissa Hall, Candace Ross, Jack Urbanek, Adina Williams, Aishwarya Agrawal, Adriana Romero-Soriano, and Michal Drozdzal. 2024.
\newblock Improving text-to-image consistency via automatic prompt optimization.
\newblock \emph{arXiv preprint arXiv:2403.17804}.

\bibitem[{Ouyang et~al.(2022)Ouyang, Wu, Jiang, Almeida, Wainwright, Mishkin, Zhang, Agarwal, Slama, Ray et~al.}]{ouyang2022training}
Long Ouyang, Jeffrey Wu, Xu~Jiang, Diogo Almeida, Carroll Wainwright, Pamela Mishkin, Chong Zhang, Sandhini Agarwal, Katarina Slama, Alex Ray, et~al. 2022.
\newblock Training language models to follow instructions with human feedback.
\newblock \emph{Advances in neural information processing systems}, 35:27730--27744.

\bibitem[{Prasad et~al.(2022)Prasad, Hase, Zhou, and Bansal}]{prasad2022grips}
Archiki Prasad, Peter Hase, Xiang Zhou, and Mohit Bansal. 2022.
\newblock Grips: Gradient-free, edit-based instruction search for prompting large language models.
\newblock \emph{arXiv preprint arXiv:2203.07281}.

\bibitem[{Pryzant et~al.(2023)Pryzant, Iter, Li, Lee, Zhu, and Zeng}]{pryzant2023automatic}
Reid Pryzant, Dan Iter, Jerry Li, Yin~Tat Lee, Chenguang Zhu, and Michael Zeng. 2023.
\newblock Automatic prompt optimization with" gradient descent" and beam search.
\newblock \emph{arXiv preprint arXiv:2305.03495}.

\bibitem[{Qin and Eisner(2021)}]{qin2021learning}
Guanghui Qin and Jason Eisner. 2021.
\newblock Learning how to ask: Querying lms with mixtures of soft prompts.
\newblock \emph{arXiv preprint arXiv:2104.06599}.

\bibitem[{Qin et~al.(2024)Qin, Chen, Feng, Wu, Zhang, Li, Li, Che, and Yu}]{qin2024large}
Libo Qin, Qiguang Chen, Xiachong Feng, Yang Wu, Yongheng Zhang, Yinghui Li, Min Li, Wanxiang Che, and Philip~S Yu. 2024.
\newblock Large language models meet nlp: A survey.
\newblock \emph{arXiv preprint arXiv:2405.12819}.

\bibitem[{Qin et~al.(2022)Qin, Chen, Xie, Li, Lou, Che, and Kan}]{qin-etal-2022-gl}
Libo Qin, Qiguang Chen, Tianbao Xie, Qixin Li, Jian-Guang Lou, Wanxiang Che, and Min-Yen Kan. 2022.
\newblock \href {https://doi.org/10.18653/v1/2022.acl-long.191} {{GL}-{CL}e{F}: A global{--}local contrastive learning framework for cross-lingual spoken language understanding}.
\newblock In \emph{Proceedings of the 60th Annual Meeting of the Association for Computational Linguistics (Volume 1: Long Papers)}, pages 2677--2686, Dublin, Ireland. Association for Computational Linguistics.

\bibitem[{Radford et~al.(2021)Radford, Kim, Hallacy, Ramesh, Goh, Agarwal, Sastry, Askell, Mishkin, Clark, Krueger, and Sutskever}]{pmlr-v139-radford21a}
Alec Radford, Jong~Wook Kim, Chris Hallacy, Aditya Ramesh, Gabriel Goh, Sandhini Agarwal, Girish Sastry, Amanda Askell, Pamela Mishkin, Jack Clark, Gretchen Krueger, and Ilya Sutskever. 2021.
\newblock \href {https://proceedings.mlr.press/v139/radford21a.html} {Learning transferable visual models from natural language supervision}.
\newblock In \emph{Proceedings of the 38th International Conference on Machine Learning}, volume 139 of \emph{Proceedings of Machine Learning Research}, pages 8748--8763. PMLR.

\bibitem[{Reynolds and McDonell(2021)}]{reynolds2021prompt}
Laria Reynolds and Kyle McDonell. 2021.
\newblock Prompt programming for large language models: Beyond the few-shot paradigm.
\newblock In \emph{Extended abstracts of the 2021 CHI conference on human factors in computing systems}, pages 1--7.

\bibitem[{Sahoo et~al.(2024)Sahoo, Singh, Saha, Jain, Mondal, and Chadha}]{sahoo2024systematic}
Pranab Sahoo, Ayush~Kumar Singh, Sriparna Saha, Vinija Jain, Samrat Mondal, and Aman Chadha. 2024.
\newblock A systematic survey of prompt engineering in large language models: Techniques and applications.
\newblock \emph{arXiv preprint arXiv:2402.07927}.

\bibitem[{Shi et~al.(2022)Shi, Suzgun, Freitag, Wang, Srivats, Vosoughi, Chung, Tay, Ruder, Zhou, Das, and Wei}]{shi2022language}
Freda Shi, Mirac Suzgun, Markus Freitag, Xuezhi Wang, Suraj Srivats, Soroush Vosoughi, Hyung~Won Chung, Yi~Tay, Sebastian Ruder, Denny Zhou, Dipanjan Das, and Jason Wei. 2022.
\newblock \href {https://arxiv.org/abs/2210.03057} {Language models are multilingual chain-of-thought reasoners}.
\newblock \emph{Preprint}, arXiv:2210.03057.

\bibitem[{Sun et~al.(2022)Sun, Shao, Qian, Huang, and Qiu}]{sun2022black}
Tianxiang Sun, Yunfan Shao, Hong Qian, Xuanjing Huang, and Xipeng Qiu. 2022.
\newblock Black-box tuning for language-model-as-a-service.
\newblock In \emph{International Conference on Machine Learning}, pages 20841--20855. PMLR.

\bibitem[{Suzgun et~al.(2022)Suzgun, Scales, Sch{\"a}rli, Gehrmann, Tay, Chung, Chowdhery, Le, Chi, Zhou, , and Wei}]{suzgun2022challenging}
Mirac Suzgun, Nathan Scales, Nathanael Sch{\"a}rli, Sebastian Gehrmann, Yi~Tay, Hyung~Won Chung, Aakanksha Chowdhery, Quoc~V Le, Ed~H Chi, Denny Zhou, , and Jason Wei. 2022.
\newblock Challenging big-bench tasks and whether chain-of-thought can solve them.
\newblock \emph{arXiv preprint arXiv:2210.09261}.

\bibitem[{Tang et~al.(2024)Tang, Wang, Zhao, Lu, Li, and Wen}]{tang2024unleashing}
Xinyu Tang, Xiaolei Wang, Wayne~Xin Zhao, Siyuan Lu, Yaliang Li, and Ji-Rong Wen. 2024.
\newblock Unleashing the potential of large language models as prompt optimizers: An analogical analysis with gradient-based model optimizers.
\newblock \emph{arXiv preprint arXiv:2402.17564}.

\bibitem[{Team et~al.(2023)Team, Anil, Borgeaud, Alayrac, Yu, Soricut, Schalkwyk, Dai, Hauth, Millican et~al.}]{team2023gemini}
Gemini Team, Rohan Anil, Sebastian Borgeaud, Jean-Baptiste Alayrac, Jiahui Yu, Radu Soricut, Johan Schalkwyk, Andrew~M Dai, Anja Hauth, Katie Millican, et~al. 2023.
\newblock Gemini: a family of highly capable multimodal models.
\newblock \emph{arXiv preprint arXiv:2312.11805}.

\bibitem[{Wang et~al.(2023)Wang, Li, Wang, Bai, Luo, Zhang, Jojic, Xing, and Hu}]{wang2023promptagent}
Xinyuan Wang, Chenxi Li, Zhen Wang, Fan Bai, Haotian Luo, Jiayou Zhang, Nebojsa Jojic, Eric~P Xing, and Zhiting Hu. 2023.
\newblock Promptagent: Strategic planning with language models enables expert-level prompt optimization.
\newblock \emph{arXiv preprint arXiv:2310.16427}.

\bibitem[{Ye et~al.(2023)Ye, Axmed, Pryzant, and Khani}]{ye2023prompt}
Qinyuan Ye, Maxamed Axmed, Reid Pryzant, and Fereshte Khani. 2023.
\newblock Prompt engineering a prompt engineer.
\newblock \emph{arXiv preprint arXiv:2311.05661}.

\bibitem[{Yuksekgonul et~al.(2024)Yuksekgonul, Bianchi, Boen, Liu, Huang, Guestrin, and Zou}]{yuksekgonul2024textgrad}
Mert Yuksekgonul, Federico Bianchi, Joseph Boen, Sheng Liu, Zhi Huang, Carlos Guestrin, and James Zou. 2024.
\newblock Textgrad: Automatic" differentiation" via text.
\newblock \emph{arXiv preprint arXiv:2406.07496}.

\bibitem[{Zamfirescu-Pereira et~al.(2023)Zamfirescu-Pereira, Wong, Hartmann, and Yang}]{zamfirescu2023johnny}
JD~Zamfirescu-Pereira, Richmond~Y Wong, Bjoern Hartmann, and Qian Yang. 2023.
\newblock Why johnny can’t prompt: how non-ai experts try (and fail) to design llm prompts.
\newblock In \emph{Proceedings of the 2023 CHI Conference on Human Factors in Computing Systems}, pages 1--21.

\bibitem[{Zhang et~al.(2022)Zhang, Wang, Zhou, Schuurmans, and Gonzalez}]{zhang2022tempera}
Tianjun Zhang, Xuezhi Wang, Denny Zhou, Dale Schuurmans, and Joseph~E Gonzalez. 2022.
\newblock Tempera: Test-time prompting via reinforcement learning.
\newblock \emph{arXiv preprint arXiv:2211.11890}.

\bibitem[{Zhao et~al.(2023)Zhao, Zhou, Li, Tang, Wang, Hou, Min, Zhang, Zhang, Dong et~al.}]{zhao2023survey}
Wayne~Xin Zhao, Kun Zhou, Junyi Li, Tianyi Tang, Xiaolei Wang, Yupeng Hou, Yingqian Min, Beichen Zhang, Junjie Zhang, Zican Dong, et~al. 2023.
\newblock A survey of large language models.
\newblock \emph{arXiv preprint arXiv:2303.18223}.

\bibitem[{Zhou et~al.(2024)Zhou, Ou, Ding, Li, Wu, Wang, Chen, Wang, Xu, Zhang et~al.}]{zhou2024symbolic}
Wangchunshu Zhou, Yixin Ou, Shengwei Ding, Long Li, Jialong Wu, Tiannan Wang, Jiamin Chen, Shuai Wang, Xiaohua Xu, Ningyu Zhang, et~al. 2024.
\newblock Symbolic learning enables self-evolving agents.
\newblock \emph{arXiv preprint arXiv:2406.18532}.

\bibitem[{Zhou et~al.(2022)Zhou, Muresanu, Han, Paster, Pitis, Chan, and Ba}]{zhou2022large}
Yongchao Zhou, Andrei~Ioan Muresanu, Ziwen Han, Keiran Paster, Silviu Pitis, Harris Chan, and Jimmy Ba. 2022.
\newblock Large language models are human-level prompt engineers.
\newblock \emph{arXiv preprint arXiv:2211.01910}.

\bibitem[{Zhu et~al.(2023)Zhu, Wang, Zhou, Wang, Chen, Wang, Yang, Ye, Zhang, Zhenqiang~Gong et~al.}]{zhu2023promptbench}
Kaijie Zhu, Jindong Wang, Jiaheng Zhou, Zichen Wang, Hao Chen, Yidong Wang, Linyi Yang, Wei Ye, Yue Zhang, Neil Zhenqiang~Gong, et~al. 2023.
\newblock Promptbench: Towards evaluating the robustness of large language models on adversarial prompts.
\newblock \emph{arXiv e-prints}, pages arXiv--2306.

\end{thebibliography}

\appendix
\onecolumn

\section*{Appendix}
\section{Detailed and Additional Experiments}
\subsection{Detailed Experiments in Best Practices}
We use GPT-4o as the backward engine and GPT-4o-mini as the forward engine. The models are trained on the BigGSM training set and tested on the BigGSM, BBH, MATH, and GSM8K test sets. The detailed results, including standard deviations, are presented below.
\begin{table}[H]
    \centering
    \footnotesize 
    \begin{tabular}{c|cccc}
    \midrule
    % \hline
    \rowcolor{gray!8}\multicolumn{5}{c}{Trainset : BigGSM} \\
    \midrule
    Method & BGSM & BBH & MATH & GSM8K \\
    \midrule
    DLPO & \textbf{60.2±2.1} & \textbf{89.7±1.5} & \underline{71.3±4.4} & \underline{93.3±3.4} \\

    TG & 55.7±2.5 & 87.3±3.7 & 69.7±3.5 & 92.7±2.1 \\

    APO & \underline{58.4±0.4} & \underline{89.0±3.4} & 59.7±9.0 & \textbf{93.4±0.4} \\

    HUMAN & 54.4 & 89.0  & \textbf{72.0} & 90.3 \\
    \midrule
    \rowcolor{gray!8}\multicolumn{5}{c}{Trainset : MGSM} \\
    \midrule
    Method & MGSM & BBH & MATH & BGSM \\
    \midrule
    DLPO & \textbf{86.7±0.2} & \textbf{89.3±3.5} & \textbf{70.8±2.5} & \textbf{55.6±1.2} \\

    TG & 61.8±17.6 & 77.7±12.0 & \underline{68.8±3.9} & 32.5±13.3 \\

    APO & \underline{82.7±3.4} & \underline{86.0±9.9} & 51.0±25.9 & \underline{47.9±12.2} \\
    \midrule
    \rowcolor{gray!8}\multicolumn{5}{c}{Trainset : BBH object counting} \\
    \midrule
    Method & BBH & GSM8K & MATH & BGSM \\
    \midrule
    DLPO & \textbf{94.2±2.0} & \textbf{93.9±0.5} & \underline{71.9±1.0} & \textbf{56.7±0.2} \\

    TG & 63.8±2.4 & 83.3±12.0 & \textbf{72.2±2.6} & 44.8±13.6 \\

    APO & \underline{90.9±2.3} & \underline{92.8±1.9} & 68.9±2.6 & \underline{55.8±0.7} \\
    \hline
    \end{tabular}
    \caption{We test various methods on the trainset of BigGSM, MGSM and BBH. Except for the HUMAN method, where the prompt is taken directly from the original article, each result represents the average and the standard deviation of three different seeds for the corresponding method. The \textbf{bold} data indicates the highest value in each column.}
    \label{tab:bstp_std}
\end{table}
\subsection{Detailed and Additional Experiments in Generalizablity}
\subsubsection{Detailed Experiments in Batch Size and Training set Size}
We use GSM8K as the prompt training set and test the performance of different batch sizes and training set sizes on the test sets of GSM8K, BigBSM, BBH, and MATH. The detailed results with standard deviations are as follows:
\begin{table}[H]
    \centering
    \small

    \begin{tabular}{l|l|cccc}
    \hline
    expname & hyper &  GSM8K & BGSM & BBH & MATH \\
    \hline
    batch size    & bs = 3    & 93.3±1.0  & \textbf{57.3±1.2}  & 79.7±12.7 & \textbf{73.9±3.1 }\\
    & bs = 6    & 93.2±2.3  & 52.7±4.2  & \textbf{85.0±13.9} & \textcolor{red}{65.6±5.7} \\
    & bs = 9    & 93.3±1.6  & 57.2±0.3  & \textcolor{red}{77.0±8.7}  & 70.2±1.6 \\
    & bs = 12  & \textbf{93.7±1.5}  & 54.8±1.6  & 79.7±13.6 & 69.6±1.8 \\
    & bs = 15  & \textcolor{red}{89.7±7.5}  & \textcolor{red}{48.4±12.4} & 80.0±12.3 & 70.8±2.5 \\
\hline
    trainset size & size = 50    & 93.2±2.3  & 52.7±4.2  & 85.0±13.9 & 65.6±5.7 \\
     & size = 100 & \textbf{94.3±0.5}  & \textbf{57.0±0.8}  & \textbf{90.3±0.6}  & \textbf{69.7±3.7} \\
    \hline

    \end{tabular}
    \caption{Accuracy comparison across different batch sizes and training set sizes. For the batch size experiments, we utilize the \textsc{Tsa}, \textsc{Tlrd}, and \textsc{Tcl} methods, which have shown promising results in previous approaches. Under the benchmark of a training set size of 50, we conduct experiments with batch sizes of 3, 6, 9, 12, and 15. For the training set size experiments, we fix the batch size at 6 and employ the \textsc{Tsa}, \textsc{Tlrd}, and \textsc{Tcl} methods to compare the performance with training set sizes of 50 and 100. All experiments are run with three different seeds, and the mean and standard deviation are calculated from the results. The \textbf{bold} data indicates the highest value in each column. The \textcolor{red}{red} data indicates the lowest value in each column.}
    \label{tab:bsts_std}
    \end{table}
\subsubsection{Additional Experiments in Generalizablity of Different LLM}
\label{appendix:gentodif_model}
We also test whether the optimized prompts obtained from specific forward (inference LLM models) and backward engines (optimizer LLM models) can generalize to other forward engines. Typically, different forward engines possess varying capabilities and excel in different domains, leading to distinct errors (loss) encountered during reasoning. This, in turn, results in different feedback (gradients), ultimately yielding varied prompts. This presents a significant challenge for generalizing optimized prompts to different forward engine.

We use the optimized prompt results obtained from training on BigGSM, with GPT-4o-mini as the forward engine and GPT-4o as the backward engine, as the system prompt for other forward LLM engines. We test these prompts in the BigGSM environment. The results for GPT-4o and GEMINI-2.0-flash are presented in table \ref{tab:gene_model}. The results demonstrate the potential of our DLPO method to generalize to other different forward engines.
\begin{table}[H]
    \centering
    \footnotesize
    \begin{tabular}{l|ccc}
    \toprule
    Method & GPT-4o-mini (original) & GPT-4o & Gemini-2.0-flash \\
    \midrule
    DLPO  &\textbf{60.2±2.1} &\textbf{77.9±1.1} & \textbf{84.0±2.7} \\
    TG  &55.7±2.5 & 75.6±3.4 &83.4±1.7    \\
    APO & 58.4±0.4 &75.5±0.2 & 83.5±1.6    \\
    HUMAN & 54.4 & 62.2 & 80.5 \\
    \bottomrule
    \end{tabular}
    \caption{We test the optimized results of DLPO, TG, APO of specific backward engine and forward engine on the trainset of BigGSM. Except for the HUMAN method, where the prompt is taken directly from the original article, each result represents the average and the standard deviation of two different optimized prompts of the corresponding method. The \textbf{bold} data indicates the highest value in each column.}
    \label{tab:gene_model}
  \end{table}
To explore the capabilities of the latest reasoning models, such as DeepSeek-R1, as the backward engine, we test three methods. The results in Table \ref{tab:gene_model_} show that our approach can effectively use DeepSeek-R1 as the backward engine and achieves the best performance among the three methods.
  \begin{table}[H]
    \centering
    \footnotesize
    \begin{tabular}{l|c}
    \toprule
    Method & DeepSeek-R1 \\
    \midrule
    DLPO  &\textbf{61.2} \\
    TG  &49.8    \\
    APO & 58.0    \\
    \bottomrule
    \end{tabular}
    \caption{We test the DeepSeek-R1 as the backward engine for DLPO, TG and APO methods. The \textbf{bold} data indicates the highest value in each column.}
    \label{tab:gene_model_}
  \end{table}
  Although DeepSeek-R1 demonstrates strong performance in optimizing prompts, it requires extended thinking time to generate feedback. Moreover, the improvement it offers over GPT-4o as the optimizer is relatively modest (1.66\%). We believe this is likely because GPT-4o has already approached the optimization limit of the prompt for the forward engine (GPT-4o-mini). Therefore, there is currently no compelling need to adopt R1-like models in practice. We plan to conduct further investigations into this in the future.

\section{Implementation Details}
\label{appen:details}
\subsection{Implementation details of Textual Learning Rate}

To address excessively large single-step updates in text optimization, we introduce the \textbf{Textual Learning Rate (\textsc{Tlr})} mechanism. This mechanism operates at the sentence level and controls the maximum number of update units allowed in a single optimization step. Each update unit corresponds to one of the following operations: (1).Adding a sentence. (2).Deleting a sentence. (3).Modifying a sentence.
% \end{itemize}
The value of \textsc{Tlr}, denoted as \( \mathcal{R} \), represents the maximum number of update units allowed in a single step. 
If the LLM intends to make more changes than \( \mathcal{R} \), we encourage it to retain impactful changes and discard non-critical ones.

Below is the original prompt that describes the concept and usage of \textsc{Tlr} for LLM prompt optimizer:

\begin{mybox}
\textbf{Instruction:}

You need to update the original variable on a sentence level, and the number of updates (including adding sentence, deleting sentence, and modifying sentence) should be limited to a specific quantity (which we call the 'learning rate').

If the learning rate is: 4, here's an example:

Initial:

<VARIABLE>

As a Math Calculator, please solve:

Required Steps:

1. Identify problem type

2. Show calculation steps

Output Format:

- Process:

- Final Result:

- Verification:

</VARIABLE>

Modified Version with exactly 4 changes:

<IMPROVED-VARIABLE>

As a reasoning Engine, please solve:  [modifying sentence]

Required Steps:

1. Identify problem type

2. Show calculation steps

3. Analyze complexity [adding sentence]

4. Assess stability [adding sentence]

Output Format:

- Process:

- Final Result:   [deleting sentence 'Verification:']

</IMPROVED-VARIABLE>

Conclusion:

1(modify) + 2(add) + 1(delete) = 4.
Your learning rate is: $\mathcal{R}$. For each optimize step, please make $\mathcal{R}$ update(s) to the original sentences and keep the other unchanged.
\label{ins:lr}
\end{mybox}

\subsection{Implementation details of Textual Dropout}
To address the high randomness of LLM updates, we introduce \textbf{Textual Dropout (\textsc{Tdo})}. 
\textsc{Tdo} requires the optimizer to randomly "drop" (skip updating) a portion of sentences during each update. 
This mechanism mitigates the risk of deleting or altering sentences with positive impact and encourages LLM to focus on updating the remaining sentences.

Formally, for \( \mathcal{S} \) total sentences and dropout rate \( p \), the number of preserved sentences \( \mathcal{K} \) is:
\begin{equation}
  \mathcal{K} = \lceil p \cdot \mathcal{S} \rceil
\end{equation}
The final set of sentences \(\mathcal{P}_{\text{new}}\) is:
\[
\mathcal{P}_{\text{new}} = \mathcal{P}_{\text{preserved}} + \mathcal{P}_{\text{updated}},
\]
where \(\mathcal{P}_{\text{preserved}}\) are the preserved sentences, and \(\mathcal{P}_{\text{updated}}\) are the updated sentences based on gradient text \(\mathcal{G}_{\text{text}}\).

Below is the algorithm design for implementing \textsc{Tdo}:

\begin{algorithm}[H]
    \caption{Textual Dropout Algorithm}
    \label{alg:tdo}
    \begin{algorithmic}[1]
    \REQUIRE Original text $T_{\text{original}}$, Dropout rate $p$
    \ENSURE Modified text $T_{\text{new}}$
    
    \STATE Split $T_{\text{original}}$ into sentences $\mathcal{S}$ using punctuation marks (e.g., ., !, ?)
    \STATE Compute the number of preserved sentences: $\mathcal{K} = \lceil p \cdot |\mathcal{S}| \rceil$
    \STATE Randomly select $\mathcal{K}$ sentences from $\mathcal{S}$ as $\mathcal{P}_{\text{preserved}}$
    \STATE Generate a T\textsc{do} instruction using $\mathcal{P}_{\text{preserved}}$

    \RETURN T\textsc{do} Instruction
    \end{algorithmic}
    \end{algorithm}
    
The selected sentences are passed to the LLM optimizer instruction, which ensures that these sentences remain unchanged during the current optimization step. The prompt is formatted as follows:

\begin{mybox}
\textbf{Instruction:}

    We have introduced a dropout mechanism. The <DROPOUT>'{sentences}'</DROPOUT> in the original variable need to remain unchanged for this optimize step. You should focus on altering the other sentences.

\end{mybox}

Here, \texttt{{sentences}} is replaced with the randomly selected sentences from the \textsc{Tdo} process.

\subsection{Implementation details of Textual Simulated Annealing}

Our \textbf{Textual Simulated Annealing (\textsc{Tsa})} scheme dynamically  combines simulated annealing principles with text optimization, allowing the prompt to escape local optima and converge to better solutions. Below is the pseudocode for implementing \textsc{Tsa}:

\begin{algorithm}[H]
    \caption{Textual Simulated Annealing Algorithm}
    \label{alg:tsa}
    \begin{algorithmic}[1]
    \REQUIRE initial prompt $x_0$, initial temperature $\mathcal{T}_0$, cooling rate $\alpha$, maximum iterations $N$
    
    \STATE initialize $x \gets x_0$, $\mathcal{T} \gets \mathcal{T}_0$
    \FOR{$i \gets 1$ to $N$}
        \STATE generate a new solution $x'$ by perturbing $x$
        \STATE compute accuracy difference: $\Delta \mathcal{E} = \mathcal{E}(x') - \mathcal{E}(x)$
        \IF{$\Delta \mathcal{E} \geq 0$}
            \STATE accept the new solution: $x \gets x'$
        \ELSE
            \STATE compute acceptance probability: $P = \exp\left(\frac{\Delta \mathcal{E}}{\mathcal{T}}\right)$
            \STATE generate a random number $r \in [0, 1]$
            \IF{$r < P$}
                \STATE accept the worse solution: $x \gets x'$
            \ENDIF
        \ENDIF
        \STATE update temperature: $\mathcal{T} \gets \alpha \cdot \mathcal{T}$
    \ENDFOR
    \STATE return $x$
    \end{algorithmic}
    \end{algorithm}
\subsubsection*{Brief Explanation}
The algorithm starts with an initial set of prompts and gradually adjusts them based on training set accuracy. The temperature parameter $\mathcal{T}$ controls the probability of accepting worse solutions, which decreases over time. This allows the model to explore the solution space early and converge to better solutions later.

\subsection{Implementation details of Textual Learning Rate Decay}
Since we emphasize the exploration of different prompting approaches at the beginning of the update process, it is crucial to encourage the LLM to make substantial modifications to the initial prompt. As the updates progress and the prompt reaches a satisfactory level, the focus shifts to fine-grained refinements, necessitating a lower learning rate for more stable and precise improvements. To achieve this, the learning rate $ \texttt{lr} $ decreases by one after each update step. For small learning rates ($ \texttt{lr} \leq 4 $), the system appends the instruction \ref{ins:lr} to the prompt, enabling fine-tuning with smaller adjustments. Conversely, for large learning rates ($ \texttt{lr} > 10 $), the system appends the instruction \ref{ins:big_lr}, which encourages more substantial modifications to explore diverse solutions. This dynamic adjustment ensures a smooth transition from exploration to refinement throughout the optimization process.

\begin{mybox}
    \textbf{Instruction:}

        In order to break away from convention, discover more creative solutions, and explore limitless possibilities, please boldly unleash your imagination based on feedback and make transformative modifications to the previous variables. Do not be confined by existing forms; courageously break the mold and experiment with entirely new combinations and ideas, as this may spark unexpected and groundbreaking outcomes.
    \label{ins:big_lr}
\end{mybox}
\subsection{Implementation Details of Textual Momentum}
To address the issue of batch-to-batch variability in optimization, we introduce \textbf{Textual Momentum (\textsc{Tm}nt)}, which leverages feedback from past updates to stabilize and refine the optimization process. 
During each update, the optimizer considers historical feedback from the last three iterations, presented using the following instruction:
\begin{mybox}
\textbf{Instruction:}

Here is the historical feedback on this variable:
<PAST-FEEDBACK>{history}</PAST-FEEDBACK>
Please analyze the main trends and patterns in the feedback across different iterations. If the feedback consistently points to similar issues or suggests insufficient modifications, it indicates that the changes made to the variable are not substantial enough. The later history feedback will be more accurate and relevant.
In such cases, please propose more significant and impactful adjustments to the variable to better address the feedback and improve its performance.

\end{mybox}
This instruction encourages the backward engine to analyze trends in historical feedback and propose meaningful adjustments when necessary. The final gradient direction is computed by incorporating past gradients with a decay factor \( \gamma \), as follows:
\begin{equation}
\mathcal{G}_{\text{final}} = \mathcal{G}_{\text{current}} + \sum_{i=1}^{3} \gamma^{i} \cdot \mathcal{G}_{\text{past}_i},
\end{equation}
where \( \mathcal{G}_{\text{current}} \) is the gradient computed from the current batch, \( \mathcal{G}_{\text{past}_i} \) represents the gradient from the \( i \)-th previous iteration, and \( \gamma \) is a decay factor (e.g., \( \gamma = 0.9 \)) that reduces the influence of older gradients. To balance recent trends without over-relying on outdated information, only the last three feedbacks are considered, and the decay factor ensures that older feedback has diminishing influence. In practice, the optimizer dynamically adjusts its reliance on past feedback based on the consistency and relevance of the trends observed in the historical data, ensuring that the optimization process is both stable and adaptive.
\subsection{Implementation Details of Textual Contrastive Learning}

To refine the optimization process, we introduce \textbf{Textual Contrastive Learning (\textsc{Tcl})}, which encourages the backward engine to distinguish high-quality (positive) prompts from low-quality (negative) ones. 
The historical prompts are split into two groups: the first half represents positive prompts, while the second half represents negative prompts. 
For positive prompts, a decreasing weight scheme prioritizes earlier high-quality samples, whereas for negative prompts, an increasing weight scheme emphasizes more recent low-quality samples. 
If a valid negative prompt is found (accuracy difference exceeds a predefined threshold), both positive and negative prompts are included in the final template; otherwise, only the positive prompt is used. 
The final gradient update is computed as:
\begin{equation}
\mathcal{G}_{\text{final}} = \mathcal{G}_{\text{current}} + \mathcal{F}_{+} - \mathcal{F}_{-},
\end{equation}
where $\mathcal{F}_{+}$ represents beneficial features from positive prompts, and $\mathcal{F}_{-}$ represents undesirable features from negative prompts.

Below is the pseudocode for implementing \textsc{Tcl}:

\begin{algorithm}[H]
    \caption{Textual Contrastive Learning Algorithm (\textsc{Tcl})}
    \label{alg:tcl}
    \begin{algorithmic}[1]
    \REQUIRE Historical Prompts ($\mathcal{H}$), Accuracy Threshold ($\tau$)
    \ENSURE Updated Prompt Template
    
    \STATE Split $\mathcal{H}$ into Positive Prompts ($\mathcal{P}$) and Negative Prompts ($\mathcal{N}$).
    
    \STATE Assign decreasing weights to $\mathcal{P}$ and increasing weights to $\mathcal{N}$.
    
    \STATE Sample two prompts from $\mathcal{P}$ and two from $\mathcal{N}$, using their respective weights.
    
    \FOR{each sampled $n \in \mathcal{N}$}
        \IF{$\max(\mathcal{P}.\text{accuracy}) - n.\text{accuracy} \geq \tau$}
            \STATE Mark $n$ as valid.
        \ENDIF
    \ENDFOR
    
    \IF{valid Negative Prompts exist}
        \STATE Generate a instruction using both $\mathcal{P}$ and valid $\mathcal{N}$.
    \ELSE
        \STATE Generate a instruction using only $\mathcal{P}$.
    \ENDIF
    
    \RETURN Generated Textual Contrastive Learning Instruction
    \end{algorithmic}
    \end{algorithm}

    The following is the instruction that uses both Positive and valid Negative Instruction for contrastive learning.
\begin{mybox}
\textbf{Instruction:}

% \label{ins:tcl}
You can learn valuable insights by comparing the good and bad variables from past data. On the training set, the better-performing variables are <Positive-VAR>$\mathcal{P}$</Positive-VAR>, while the poorer-performing variables are <Negative-VAR>$\mathcal{N}$</Negative-VAR>. To improve your variable, focus on adopting the unique features that contribute to the success of the better variables and eliminate the unique features associated with the poorer variables. This approach will help enhance performance and avoid repeating past mistakes.

\end{mybox}

The following is the instruction that only encourages imitation of positive prompts.
\begin{mybox}
\textbf{Instruction:}

% \label{ins:imi}
You can gain valuable insights by analyzing high-performing variables from historical data. On the training set, the top-performing variables are <<Positive-VAR>$\mathcal{P}$</Positive-VAR>.

\end{mybox}

It is worth noting that if we use both T\textsc{sa} and T\textsc{cl}, the solutions rejected by T\textsc{sa} are actually tested for accuracy on the train set, and as a result, we will also include this result in our historical prompts for T\textsc{cl}.

\subsection{Implementation Details of Textual Regularization}
To enhance the generalization capability of our approach, we incorporate regularization techniques into the text optimization process. Specifically, we apply L2 regularization to simplify overly complex sentences and L1 regularization to eliminate irrelevant ones. In the context of text gradients, each sentence is treated as a distinct feature.

The implementation involves the following steps:

\begin{itemize}
    \item \textbf{L2 Regularization (Simplification)}: 
    We encourage simplifying individual sentences by reducing their complexity while preserving their core meaning. This is achieved through the following instruction:
    \begin{mybox}
    \textbf{Instruction:}
    
    Please simplify the overly complex and lengthy sentences in the variable. Ensure the output is concise, easy to understand, and suitable for a general audience.
    \end{mybox}

    \item \textbf{L1 Regularization (Sparsity)}: 
    We promote sparsity by removing sentences that are deemed irrelevant or detrimental to the overall meaning. The corresponding instruction is as follows:
    \begin{mybox}
    \textbf{Instruction:}

    If you are certain that a particular sentence in the variable has no impact on the overall meaning or purpose or has a negative effect, please delete that sentence. However, if you believe that all sentences are useful and contribute to the overall meaning, then retain all sentences. Ensure that the final variable maintains clarity, coherence, and relevance.
    \end{mybox}
\end{itemize}

This approach ensures that the optimized text is both concise and meaningful, striking a balance between simplicity and relevance.
\section{Experimental Setup}
Apart from the additional experiments provided in the appendix \ref{appendix:gentodif_model}, all experiments in the main text consistently use GPT-4o as the Backward engine and GPT-4o-mini as the Forward engine.
\subsection{Benchmark Description}
\label{appendix:benchmark}
In our experiments, we utilize the below benchmarks to evaluate the effectiveness of our proposed methods:
\begin{itemize}
    \item GSM8K: A collection of 8.5K high-quality grade school math word problems designed to assess language models' multi-step mathematical reasoning capabilities. 
    \item MATH: A dataset covering a wide range of topics to test advanced mathematical reasoning.
    \item BigGSM: An extension of the GSM8K dataset, including more complex and diverse grade school math problems to further challenge model performance.
    \item BigBenchHard (BBH): A subset of the BIG-Bench benchmark focusing on particularly challenging tasks across various domains, designed to push the limits of language model understanding and reasoning.
    \item MGSM: A multilingual version of the GSM8K dataset, containing grade school math problems translated into multiple languages to evaluate models' cross-lingual mathematical reasoning abilities.
\end{itemize}

\subsection{Experimental Setup in Exploration for T\textsc{M}nt and T\textsc{cl}}
In the Exploration for Efficiency section, we employ Textual Contrastive Learning and Textual Momentum methods to expedite convergence speed.
We define convergence as the validation set accuracy of the prompt reaching a certain threshold and maintaining this level for three consecutive update steps (the threshold is set at 0.5 for BigGSM and 0.8 for MGSM).
We showcase the average and standard deviation of the convergence steps using five different seeds.
\subsection{Experimental Setup in Best Practices Section}
\label{appendix:expset}
In the context of best practices, our DLPO, classical TG, and APO methods are assessed on the training datasets of BigGSM, MGSM, and BBH. 
The final step results are chosen for evaluation. 
With the exception of the HUMAN technique, which relied on prompts extracted directly from the source articles, each reported performance metric reflects the average of three distinct seed values per method. 
The hyperparameters utilize across the three distinct training datasets are detailed in Table \ref{tab:hy_bp}.
\begin{table}[H]
    \centering
    \begin{tabular}{l|lll}
    \hline
    \textbf{Argument} & \textbf{BigGSM}&\textbf{MGSM}&\textbf{BBH}\\
    \hline
    Trainset Size & 200& 50 &50\\
    Backward Engine & GPT-4o& GPT-4o& GPT-4o\\
    Forward Engine & GPT-4o-mini& GPT-4o-mini& GPT-4o-mini\\
    Batch Size & 3& 2 &3\\
    Epochs & 1& 1 &2\\
    \textsc{Tlr} & False& False & False\\
    \textsc{Tdo} & False& False & False\\
    \textsc{Tr}egu & True& True & True\\
    \textsc{Tcl} & True & True & True\\
    \textsc{Tsa} & True & True & True\\
    \textsc{Tm}nt & False & False & False\\
    \textsc{Tlrd} & 60 & 25 & 30\\
    \hline
    \end{tabular}
    \caption{Hyperparameter settings for different trainsets in Best Practices section}
    \label{tab:hy_bp}
\end{table}

\section{Optimized Results on Benchmark Datasets}
In this section, we present the three original prompts before averaging in the best practices section, each generated with the same hyperparameters but different random seeds. Additionally, the updated results are influenced by the uncertainty of the API interface. Each prompt represents the last prompt obtained throughout the entire updating process. It is worth noting that our prompt will be sent into the API as a system prompt, rather than directly concatenated with the question.
\subsection{Results on BigGSM Dataset}
\begin{mybox}
    \textbf{Optimized Prompt 1:}

Provide a clear, detailed step-by-step calculation for the reasoning question. Emphasize accuracy by verifying calculations at each step, especially in multiplication, and reassessing both intermediate and final results. Outline all assumptions and define variables explicitly to reduce ambiguity. Ensure transparency by showing intermediate steps, enabling error detection and understanding. After obtaining the solution, cross-check and simplify calculations for consistency. Evaluate the plausibility of results in a real-world context to identify any anomalies. Present the final numerical answer clearly, with consistent units where applicable. Remain cautious of overconfidence and express uncertainty if appropriate.
\end{mybox}

\begin{mybox}
    \textbf{Optimized Prompt 2:}

For reasoning questions, provide clear, detailed, and contextually relevant responses. To ensure numerical accuracy and clarity:
    
1. **Context Recognition**: Define the scenario and identify key parameters. Use suitable formulas and units for the context.

2. **Calculation Transparency**: Clearly show each calculation step and verify accuracy with known results or alternative methods.

3. **Error Checks**: Detect and correct errors in real-time.

4. **Format Flexibility**: Use diverse examples to demonstrate adaptability to different numerical formats, like with and without commas.

5. **Consistent Formatting**: Use standard numerical presentations, such as "1,234," for readability unless specified otherwise.

6. **Format Validation**: Cross-check numerical formats with trusted sources for compliance.

7. **Feedback Use**: Continuously refine formats using feedback to enhance precision.

8. **Final Check**: Ensure final outputs meet expected formats and make adjustments as needed.

Final Output: Present the final number clearly on a new line, typically formatted as "9876" for easy identification.
\end{mybox}

\begin{mybox}
    \textbf{Optimized Prompt 3:}

Deliver precise and effective answers to reasoning questions by implementing these strategies:

1. **Clarify the Question:** Restate the question to ensure complete understanding.

2. **Simplify Calculations:** Break down arithmetic tasks into smaller steps and verify each.

3. **Log Steps Clearly:** Document each calculation step methodically, minimizing rounding errors until the final result.

4. **Verify Results:** Cross-check numerical results using alternative methods or sources.

5. **Use Examples:** Apply relatable examples for better understanding.

6. **Restate Instructions:** Clearly articulate all numerical relationships to avoid omissions.

7. **Detect Errors:** Use checks for inconsistencies, such as unexpected numerical values, to identify errors.

8. **Intermediate Checks:** Implement sanity checkpoints or intermediate results for large calculations.

9. **Iterate and Improve:** Continuously refine responses through feedback and iterative checks.

10. **Prioritize Complexity:** Address the most complex or error-prone aspects first.

These strategies focus on accurate data extraction, direct calculations, contextual interpretation, and consistent cross-verification, enhancing clarity, precision, and adaptability in reasoning tasks.
\end{mybox}
\subsection{Results on MGSM Dataset}
\begin{mybox}
    \textbf{Optimized Prompt 1:}

    1. Clearly state the main answer in a simple numeric form: "Answer: [numeric value]".

2. Use the following strategies for clarity and precision:

- Utilize bullet points for concise step-by-step guidance.

- Include only the necessary context for calculations.

- Prioritize numerical precision and verify accuracy.

- Define variable roles consistently and standardly.

- Implement checks to ensure alignment with expected results and avoid common errors.

- Manage units and contexts effectively.

- Conclude with a concise summary of key supporting points.
\end{mybox}
\begin{mybox}
    \textbf{Optimized Prompt 2:}

Break down the reasoning process into clear steps, ensuring correct unit handling. Verify unit consistency and apply necessary conversions. Use a structured format to improve clarity.

1.**Explicit Assumptions**: Clearly state any assumptions, especially those related to units or context. Make these explicit to avoid misunderstandings.

2.**Mathematical Verification**: Verify all calculations and unit conversions explicitly. Check each step for accuracy, using past examples for cross-reference.

3.**Diverse Reasoning Methods**: Apply both deductive and analogical reasoning. Ensure the purpose of each step is clear, including unit handling.

4.**Clear Language**: Use precise language to avoid ambiguities, especially regarding units and quantities. Make each step understandable.

5.**Intermediate Steps**: Break down reasoning into intermediate steps to show logical flow and the significance of each part.

6.**Feedback Anticipation**: Anticipate potential feedback or errors, especially with unit conversions. Adapt approaches to address these proactively.

7.**Confidence Levels**: Express uncertainties and confidence levels, focusing on conclusions related to unit handling.

8.**Verified Final Answer**: Provide a precise final answer, checking for accuracy in unit-related aspects. Highlight the final answer for clarity.

9.**Exact Matching and Clarification**: Ensure outputs match the correct answers exactly and include contextual clarifications, focusing on units.

10.**Self-Evaluation**: Evaluate responses for unit accuracy, clarity, and brevity. Address key points thoroughly.

Maintain consistent response quality by integrating feedback to enhance precision and relevance. Use adaptive learning to focus on historical errors and improve unit handling continuously.
\end{mybox}

\begin{mybox}
    \textbf{Optimized Prompt 3:}

    1. Provide a concise numerical answer to the reasoning question through these steps:

   - Analyze the query to identify essential data points directly relevant to the answer.

   - Use a step-by-step logical computation approach, clearly linking each step to the query's context.

   - Consistently use terminology from the input data to prevent ambiguity.
   - Verify calculations with provided data, providing brief justifications for each key operation.

   - Maintain a consistent numerical format to prevent parsing errors and adhere to local and cultural conventions if specified.

   - Tailor your explanation to the user's expertise: Offer detailed steps for beginners and concise steps for advanced users.

   - Present the final numerical answer in square brackets, e.g., [42], in its simplest form, ensuring consistent units unless additional formatting is specified.

   - Guarantee the output is a single, precise numerical value, fulfilling evaluation metrics and ensuring logical coherence.
\end{mybox}
\subsection{Results on BBH Object Counting Trainset}

\begin{mybox}
    \textbf{Optimized Prompt 1:}

You will address quantitative reasoning and arithmetic questions with precision and clarity. Ensure arithmetic accuracy by verifying calculations and cross-referencing results with known benchmarks. Break down calculations into clear steps for transparency and unambiguous answers. Contextualize responses by considering relevant prior information.

Implement robust error detection and handling, including recalculations and user clarification requests when discrepancies arise. Verify counts explicitly by listing items and confirming totals. Use clear language to confirm counts and details, highlighting important numerical information.

Adopt a strategy of self-evaluation, focusing on accuracy and relevance before finalizing responses. Avoid unnecessary elaboration unless requested. Develop a systematic approach for identifying and correcting errors, including a secondary verification mechanism for counts.

Deliver concise responses to boost clarity and user satisfaction, especially when detailed explanations are unnecessary. Address ambiguous queries by clarifying assumptions or soliciting additional information. Ensure uniform formatting to emphasize critical numerical data. Continuously adapt responses to enhance user comprehension.

Establish a feedback loop to learn from past interactions, refining response strategies to align with expected outcomes. Use digit form for numerical data to align with ground truth expectations. Format responses for easy parsing by automated systems, ensuring consistency with the ground truth. Implement a mechanism for continuous improvement of numerical accuracy based on feedback.

Use examples from training data to guide response formatting and accuracy. Ensure responses logically follow from the query and context, directly addressing the question. Incorporate a mechanism for detecting and correcting errors by flagging inconsistencies and prompting user verification. Use examples that demonstrate clear enumeration and counting strategies to align responses with best practices.
\end{mybox}

\begin{mybox}
    \textbf{Optimized Prompt 2:}

    You will answer reasoning questions with a focus on tasks that may involve counting. Follow these streamlined guidelines to ensure clarity, precision, and efficiency:

    1. **Task Recognition**: Identify if the question involves counting and provide a precise numerical answer.
    
    2. **Direct Instructions**: Recognize lists or enumerations as signals for counting tasks and deliver a straightforward numerical count.
    
    3. **Contextual Awareness**: Be aware of contextual cues that suggest a counting task and ensure your response is relevant.
    
    4. **Logical Response Structure**: Clearly list relevant items, then provide a total count, using simple formatting to distinguish steps.
    
    5. **Verification and Accuracy**: Verify your numerical responses for accuracy and consistency with the listed items.
    
    6. **Example Utilization**: Use concise examples to illustrate clarity and consistency in handling counting tasks.
    
    7. **Efficiency Feedback Loop**: Internally verify your responses to maintain accuracy and logical flow.
    
    By following these strategies, you will deliver concise, precise, and contextually appropriate answers to reasoning questions, optimizing for runtime efficiency.
\end{mybox}

\begin{mybox}
    \textbf{Optimized Prompt 3:}

Begin each response by restating the task briefly for clarity. Provide the main answer upfront, especially for numerical queries, and include explanations only if they add significant value. Use bullet points or lists for complex information to enhance clarity. Avoid redundancy and unnecessary details, ensuring responses are concise and self-contained. Address ambiguities by seeking clarification or clearly stating assumptions. Maintain a conversational tone and anticipate follow-up questions to improve interaction. Present reasoning in logical steps and offer detailed breakdowns of calculations when necessary. Implement verification mechanisms, such as checklists or summary statements, to confirm accuracy. Ensure numerical operations align with expected results and conclude with a verification statement. Consistently align responses with the original query, rephrasing if needed for comprehension. Learn from past inaccuracies to improve future responses. Consider edge cases like zero quantities or duplicates and adjust calculations accordingly. Use a feedback loop for clarification, prompting follow-up questions or stating assumptions if the task is unclear. Integrate numerical examples and encourage step-by-step calculations to ensure consistency and verify results by checking individual counts against the total.
\end{mybox}
\label{sec:appendix}

\end{document}